\documentclass[10pt,twocolumn,letterpaper]{article}

\usepackage[pagenumbers]{cvpr} 
\usepackage[accsupp]{axessibility}
\usepackage{xcolor}
\usepackage{hhline}
\usepackage{boldline}
\usepackage{colortbl}
\usepackage{amsfonts}
\usepackage{pifont}
\usepackage{amsmath}
\usepackage{booktabs}
\usepackage{multirow}
\usepackage{array}
\definecolor{lightred}{RGB}{255,153,153} 

\usepackage{algorithm}
\usepackage{algorithmic}
\usepackage{subcaption}
\usepackage{enumitem}
\usepackage{tabularx}

\definecolor{cvprblue}{rgb}{0.21,0.49,0.74}
\usepackage[pagebackref,breaklinks,colorlinks,allcolors=cvprblue]{hyperref}

\newcommand*{\affaddr}[1]{#1} 
\newcommand*{\affmark}[1][*]{\textsuperscript{#1}}

\def\confName{CVPR}
\def\confYear{2025}

\title{FinePhys: Fine-grained Human Action Generation by Explicitly Incorporating Physical Laws for Effective Skeletal Guidance}

\author{%
\textbf{Dian Shao}\affmark[1]\thanks{Corresponding Author}~~~
\textbf{Mingfei Shi}\affmark[1]~~~
\textbf{Shengda Xu}\affmark[2]~~~
\textbf{Haodong Chen}\affmark[3]~~~
\textbf{Yongle Huang}\affmark[3]~~~
\textbf{Binglu Wang}\affmark[4]\\
\affaddr{\affmark[1]Unmanned System Research Institute, Northwestern Polytechnical University, Xi'an, China}\\
\affaddr{\affmark[2]School of Software, Northwestern Polytechnical University, Xi'an, China}\\
\affaddr{\affmark[3]School of Automation, Northwestern Polytechnical University, Xi'an, China}\\
\affaddr{\affmark[4]School of Astronautics, Northwestern Polytechnical University, Xi'an, China}\\
}
\begin{document}
\maketitle
\begin{abstract}
Despite significant advances in video generation, synthesizing physically plausible human actions remains a persistent challenge, particularly in modeling fine-grained semantics and complex temporal dynamics.
For instance, generating gymnastics routines such as \textit{“switch leap with 0.5 turn”} poses substantial difficulties for current methods,  often yielding unsatisfactory results. To bridge this gap, we propose \textbf{FinePhys}, a \textbf{Fine}-grained human action generation framework that incorporates \textbf{Phys}ics to obtain effective skeletal guidance. Specifically, FinePhys first estimates 2D poses in an online manner and then performs 2D-to-3D dimension lifting via in-context learning. To mitigate the instability and limited interpretability of purely data-driven 3D poses, we further introduce a physics-based motion re-estimation module governed by Euler-Lagrange equations, calculating joint accelerations via bidirectional temporal updating. The physically predicted 3D poses are then fused with data-driven ones, offering multi-scale 2D heatmap guidance for the diffusion process. Evaluated on three fine-grained action subsets from FineGym (FX-JUMP, FX-TURN, and FX-SALTO), FinePhys significantly outperforms competitive baselines. Comprehensive qualitative results further demonstrate FinePhys's ability to generate more natural and plausible fine-grained human actions. 
Project Page: \textcolor{lightred}{\href{https://smartdianlab.github.io/projects-FinePhys/}{\textbf{FinePhys Webpage}}}.

\end{abstract}    
\vspace{-0.4em}

\section{Introduction}\label{introduction}
\begin{figure}[t]
    \centering
    \vspace{-0.5em}
    \includegraphics[width=1.0\linewidth]{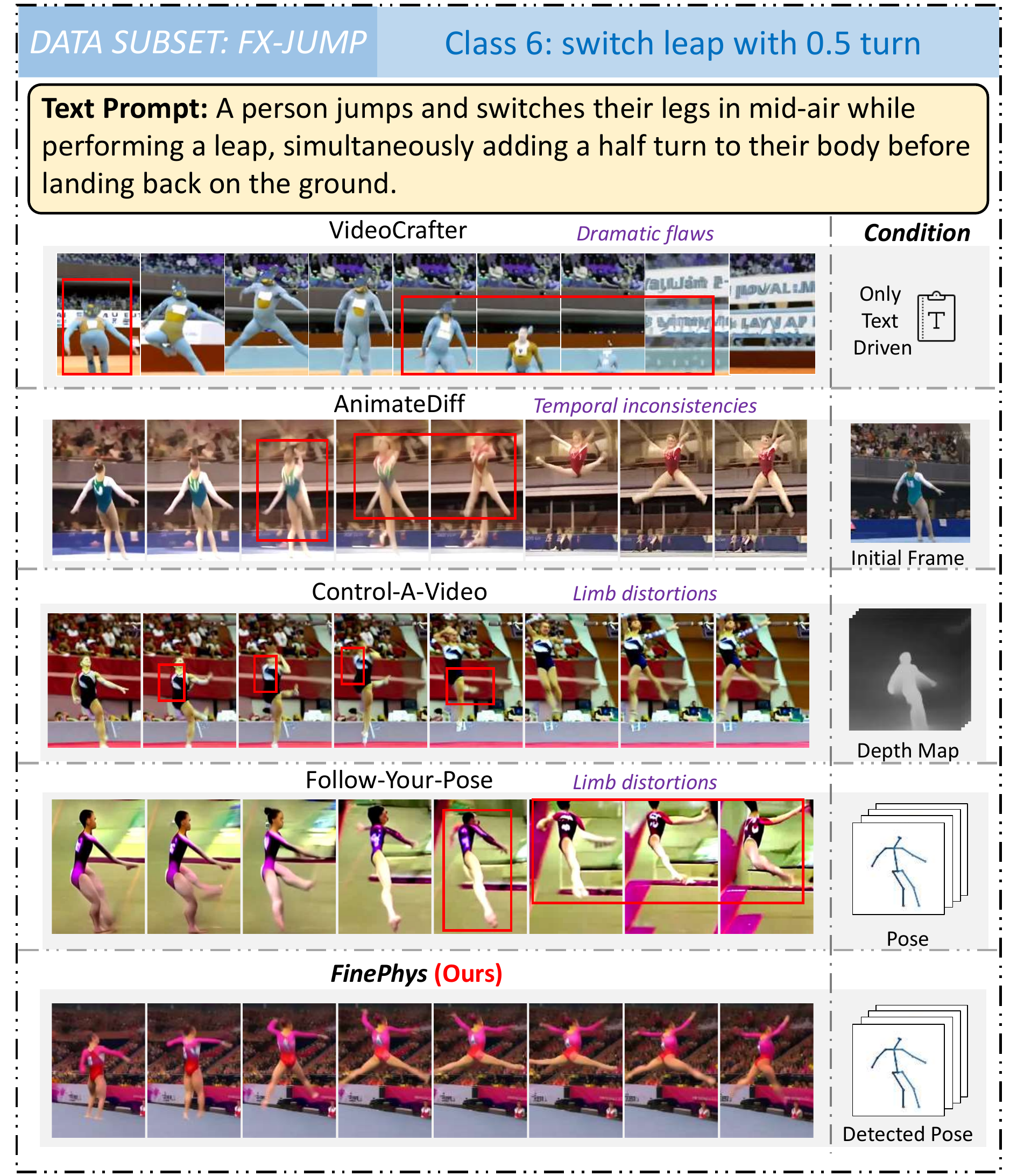}
    \vspace{-1.9em}
    \caption{Video generation results for fine-grained human action \textit{``split leap with 1 turn"}. Our \textbf{FinePhys} demonstrates superior performance in generating physically plausible fine-grained human actions, while SOTA methods exhibit significant issues, including severe temporal inconsistencies~\cite{guo2023animatediff}, noticeable limb distortions~\cite{ma2024follow}, and character anomalies~\cite{chen2023videocrafter1}.
    }
    \label{fig:instance}
    \vspace{-1.4em}
\end{figure}


The rapid evolution of generative models, particularly diffusion models~\cite{ho2020denoising,rombach2022high}, 
has significantly advanced progress in video generation. 
However, new challenges have emerged, as modeling temporal variations—such as camera motions~\cite{wang2024motionctrl}, background changes~\cite{khachatryan2023text2video}, and character movements~\cite{ma2024follow}—remains inherently difficult.
These challenges are especially pronounced in generating human actions, often leading to unnatural and inconsistent results~\cite{guo2023animatediff, esser2023structure,videoworldsimulators2024}. 
Spatially, the human body exhibits strong structural coherence, which often causes models to generate abnormal anatomical features~\cite{fang2024humanrefiner}. 
Temporally, motions must obey kinematic laws, yet recent studies~\cite{kang2024farvideogenerationworld} show that
even state-of-the-art generative models fail to preserve fundamental physical principles such as Newton's laws of motion.

In this work, we focus on an even more challenging task: \textit{generating fine-grained human actions} involving large body deformations and significant temporal changes. For example, when attempting to generate gymnastics actions, \textit{e.g.}, \textit{``Split leap with 1 turn"}, existing state-of-the-art methods fail to provide satisfactory results (see Fig.~\ref{fig:instance}). The biomechanical structure of human bodies in these cases is poorly preserved, let alone the plausibility of motion dynamics.

To address these challenges, we introduce FinePhys, a physics-aware framework for fine-grained human action generation, as shown in Fig.~\ref{fig:2}.
Specifically, besides textual input, FinePhys first extracts online 2D poses from input videos, serving as a compact prior for the biophysical structure.
Then, using the newly proposed in-context learning technique, the 2D poses are lifted to 3D poses to enhance spatial perception.
However, such purely data-driven 3D poses could ignore physical laws of motion, thus we propose a \textit{PhysNet} module that enforces Newtonian mechanics through Euler-Lagrange equations for rigid-body dynamics. 
This module bidirectionally re-estimates joint positions by modeling second-order kinematics (accelerations), yielding physics-refined 3D pose sequences.  
Finally, both data-driven and physically predicted 3D poses are fused, projected to 2D, and further encoded to provide multi-scale heatmaps, guiding the 3D-UNet denoising process.

The key question is \textit{how to coherently incorporate physical laws} into the learning process. 
Traditionally, there are three strategies~\cite{ banerjee2024physics, yuan2023physdiff}: \textit{observational bias} (via data), \textit{inductive bias} (via networks), and \textit{learning bias} (via losses). FinePhys integrates physics through all these aspects. 
Specifically,
\ding{182} For \textit{observational bias}, We include pose as an additional modality to encode biophysical layouts and utilize in-context learning for 2D-to-3D lifting, where mean 3D poses from existing datasets are used as pseudo-3D references. 
\ding{183} To encode stronger \textit{inductive biases} into FinePhys~\cite{kadkhodaiegeneralization}, we instantiate Lagrangian rigid body dynamics through fully differentiable neural network modules, whose output are parameters in the Euler-Lagrangian equation.
\ding{184} For \textit{learning bias}, we implement loss functions that adhere to the underlying physical processes.
The main contributions are summarized as follows:
\begin{itemize}[leftmargin=*]
    \item We develop \textit{FinePhys}, a novel framework for fine-grained human action video generation, which employs skeletal data as structural priors and \textit{explicitly} encodes Lagrange Mechanics via dedicated network modules; 
  \item  FinePhys incorporates physics into the generation process through multiple strategies, including the \textit{observational bias} (2D-to-3D dimension lifting), \textit{inductive bias} (Phys-Net for parameter estimation in the Euler-Lagrangian equation), and \textit{learning bias} (corresponding losses);
  \item Extensive experiments on fine-grained action subsets demonstrate that FinePhys significantly outperforms various baselines in producing more natural and physically plausible results.
\end{itemize}

\section{Related Work}
\makeatletter
\renewcommand\subsubsection{\@startsection{subsubsection}{3}{\z@}%
                                     {-3.25ex\@plus -1ex \@minus .2ex}%
                                     {-1em}%
                                     {\normalfont\normalsize\bfseries}}
\makeatother
\vspace{-0.4em}
\subsubsection*{Video Generation with Diverse Guidance.} 
Video generation has been significantly advanced by the development of deep generative models \cite{sauer2023stylegan,esser2021taming,ramesh2021zero}, particularly diffusion models \cite {ho2020denoising, song2020score}. The denoising process in diffusion models can be performed either directly in the pixel space or within a lower-dimensional latent space \cite{ma2024latte,rombach2022high,yu2023video,an2023latent,gupta2024photorealistic}, 
and FinePhys adopts the latter for efficiency.
Early approaches in video generation extend the successful text-to-image (T2I) \cite{ramesh2022hierarchical, rombach2022high, nichol2021glide, saharia2022photorealistic, raffel2020exploring, radford2021learning} to text-to-video (T2V) generation \cite{chen2025temporal, ma2024follow, chen2023control, chen2024omnicreator}. 
Although vivid frames could be produced, relying solely on textual guidance offers limited control over both spatial layouts and temporal dynamics.

Recent methods have incorporated diverse forms of guidance and additional modalities to enhance control and realism in video generation, which roughly fall into two aspects, \textit{i.e.}, appearance \cite{guo2023animatediff, he2023animate,kwon2024harivo, yan2024dialoguenerf} and structure \cite{ma2024follow, chen2023control, zhao2025motiondirector}.
Examples from the former include generating videos conditioned on an image~\cite{kandala2024pix2gif} (\textit{e.g.}, the first frame~\cite{guo2025sparsectrl, guo2023animatediff}, the last frame~\cite{oh2024mevg}) or enabling appearance customization into pre-trained T2I~\cite{gal2022image, Lu_2023_CVPR}.
The latter tries to utilize more structural guidance (\textit{e.g.}, depth~\cite{lapid2023gd,liang2024movideo}, skeleton~\cite{chen2023control, ma2024follow,ju2023humansd,zhao2024magdiff}, edges~\cite{chen2023control}, optical flows~\cite{ni2023conditional,liang2024movideo}, and trajectory~\cite{yin2023dragnuwa}), combined with ad-hoc feature encoders~\cite{zhang2023adding, mou2024t2i} to guide the generation process.
To distinguish, our FinePhys online estimates 2D pose and transforms it into 3D skeletons for enhanced spatial guidance, and incorporates an awareness of motion laws through the proposed PhysNet module. 

\vspace{-1.3em}
\subsubsection*{Physics-informed Action/Motion Modeling.}
To achieve a more realistic and reasonable motion modeling, several methods have emphasized the utilization of physics. 
Some works take advantage of Physics engines~\cite{shimada2021neural,yuan2021simpoe,gartner2022differentiable,huang2022neural,gartner2022trajectory}.
PhysGen~\cite{liu2024physgen} employs rigid-body physics simulations to convert a single image and input forces into realistic videos, demonstrating the possibility of reasoning physical parameters from visual data.
PhysDiff~\cite{yuan2023physdiff} also performs motion imitation within a physics simulator by embedding a physics-based projection module that iteratively guides the diffusion process. However, PhysDiff addresses only global artifacts such as penetration and floating, neglecting fine-grained human joint details.
LARP~\cite{andriluka2024learned} propose a novel neural network as an alternative to traditional physics simulators to facilitate human action understanding from videos.

Additionally, direct application of physical equations has also proved to be effective~\cite{xie2021physics,zhang2022pimnet,zhang2024physics}.
PIMNet~\cite{zhang2022pimnet} calculates human dynamic equations for future motions, but the joint state can be directly obtained from MoCap data, while our FinePhys estimates each state solely based on video input.
Recently, PhysPT~\cite{zhang2024physpt} has been proposed to estimate human dynamics from monocular videos based on SMPL representation. 
 However, it incorporates physics into the training of neural networks through Lagrangian losses, whereas our FinePhys explicitly estimates physical parameters of Euler-Lagrange equations (EL-Eq.).
 Furthermore, PhysMoP~\cite{zhang2024incorporating}, designed for motion prediction, also relies on EL-Eq. but focuses on predicting future SMPL pose parameters based on previous ones, which is a straightforward process.
 In contrast, FinePhys tackles a more challenging task that involves modality transformation, dimension lifting, and visual content generation.
Compared with previous works, FinePhys: 
(1) focuses on the extremely challenging task of generating fine-grained human action videos;
(2) uses monocular videos as input, online estimates 2D poses, and transforms them into 3D through in-context learning; 
(3) \textit{explicitly} instantiates EL-Eq. through the PhysNet module and calculates temporal variations of each joint \textit{bidirectionally} without relying on simulators.

\begin{figure*}
    \centering
    \vspace{-0.7em}
    \includegraphics[width=1.0\linewidth]{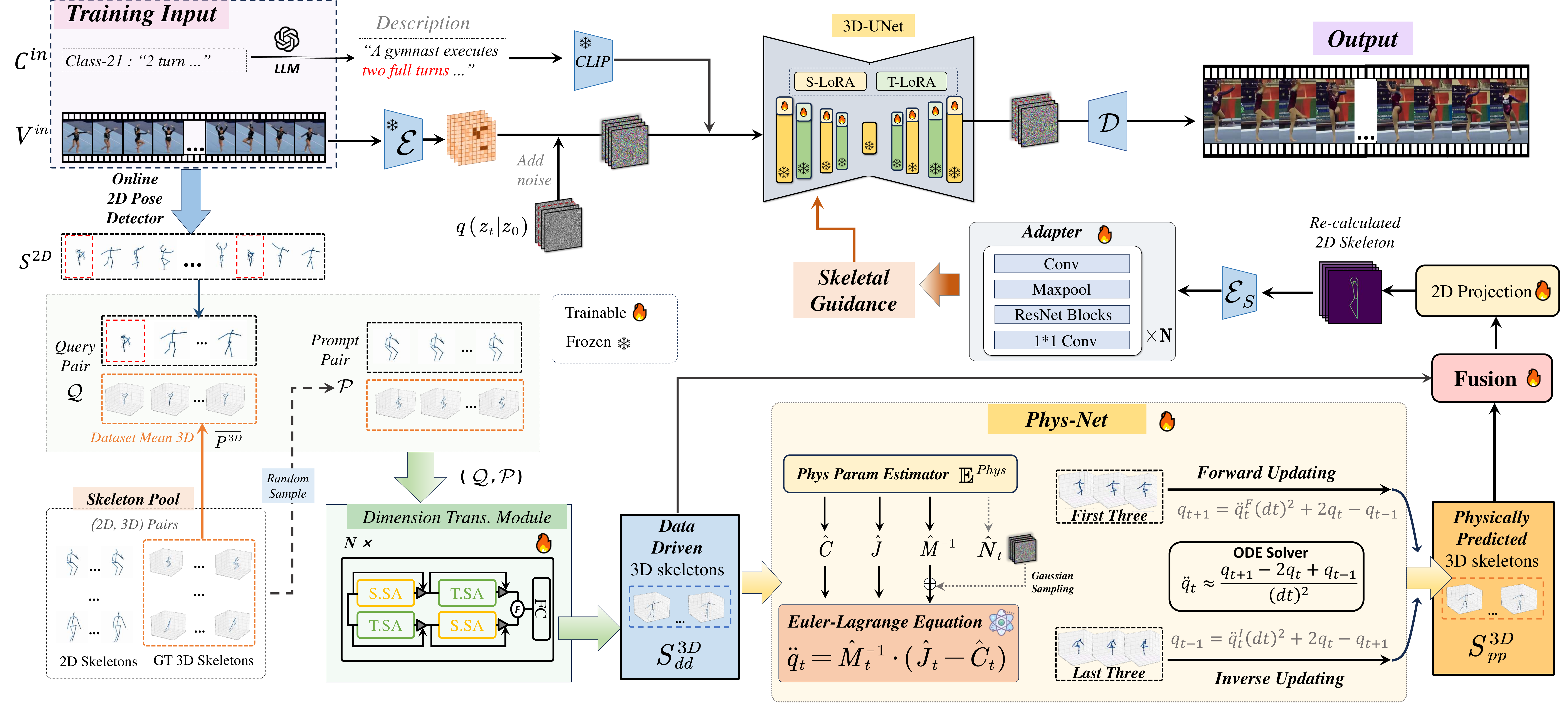}
    \vspace{-1.9em}
    \caption{\textbf{Overview of Finephys.}
    FinePhys addresses the challenging task of generating fine-grained human action videos by explicitly incorporating physical equations exploiting pose modality.
    The pipeline begins with online extracting 2D poses, then transforms them into 3D using an in-context learning module, achieving the data-driven 3D skeleton sequence $S^{3D}_{dd}$.  
    To incorporate the physical laws of motion, we introduce a Phys-Net module to re-estimate the 3D positions of each human joint by accounting for second-order temporal variations (\textit{i.e.}, accelerations) in both forward and reverse directions, yielding physically predicted 3D poses $S^{3D}_{pp}$.
    Subsequently, $S^{3D}_{dd}$ and $S^{3D}_{pp}$ are fused, projected back into 2D space, encoded into multi-scale latent maps, and integrated into 3D-UNets to guide the denoising process.
    }
    \label{fig:2}
    \vspace{-0.4em}
\end{figure*}

\vspace{-0.4em}
\section{Methodology}
\vspace{-0.2em}
\subsection{Preliminaries}
\vspace{-0.2em}
\noindent\ding{118} \textbf{Latent Diffusion Models (LDMs)} \cite{rombach2022high} are widely used for generating visual content including images and videos. A general pipeline is to use a pre-trained autoencoder $\mathcal{E}(\cdot)$ to compress high-dimension information $x$ into low-dimension latent representation $z$, \textit{i.e.}, $\mathcal{E}(x) = z$, and then modeled with trainable DDPM~\cite{ho2020denoising}:
{{\setlength\abovedisplayskip{3pt}
\setlength\belowdisplayskip{3pt}
\begin{align}
    \ell = \mathbb{E}_{\mathcal{E}(x), y, \epsilon\sim\mathcal{N}(0, 1), t} [\|\epsilon - \epsilon_\theta(z_t, t, \tau_\theta(y)\|_2^2],
\end{align}}

\noindent where $\tau_\theta(\cdot)$ denotes the condition encoder (\textit{e.g.}, CLIP text encoder~\cite{radford2021learning}). The generated results are obtained via denoising in the latent space with condition guidance encoded.

\noindent\ding{118} \noindent\textbf{Physics of Rigid-body Dynamics} could be modeled with diverse physical approaches (\textit{e.g.}, Newtonian, Lagrangian, or Hamiltonian), and they result in the equivalent sets of equations~\cite{murray2017mathematical}. 
Among these, the Euler–Lagrange Equations (EL-Eq.) are widely used to predict the dynamics of rigid bodies.
Assume $X$ as a generalized coordinate system and $q(t)\in \mathbb{X}$ as a position function of time, $L$ is the Lagrangian.
Given $q \in \mathbb{P}(a, b, x_a, x_b)$ satisfying $q:[a, b] \to X$ with $q(a) = x_a, q(b) = x_b$, the EL-Eq. can be defined as:
{\setlength\abovedisplayskip{3pt}
\setlength\belowdisplayskip{3pt}
\begin{align}
    \frac{\partial L}{\partial q^i}(t, q(t), \dot{q}(t)) - \frac{d}{dt}\frac{\partial L}{\partial \dot{q}^i} (t, q(t), \dot{q}(t)) = 0,
\end{align}}

\noindent where $\dot{q}$ and $\ddot{q}$ represent the velocities and accelerations of the joint respectively.
For the kinematics of the full-body human model, the EL-Eq. can be converted into\footnote{Details can be found in the Supplementary, as described in~\cite{liu2012quick}.} 
{\setlength\abovedisplayskip{3pt}
\setlength\belowdisplayskip{3pt}
\begin{align}
    M(q) \ddot{q} = J(q, \dot{q}) - C (q, \dot{q}),
\end{align}}

\noindent where $M$ is the generalized inertia matrix including body mass and other inertia terms; $J$ is a vector of generalized forces acting on the human body, and $C$ denotes all other terms to enforce joint constraints.

\vspace{-0.2em}
\subsection{Overview}
\vspace{-0.2em}
\subsubsection*{\ding{113} Task Definition and Problem Setting.} In this work, we focus on a novel and challenging task of generating fine-grained human action videos.
Specifically, the inputs during training are two folds: 
(1) $V^{in}=\{f_i\}_{i=1}^{T}$: a set of $T$ sampled frames from the entire video; and
(2) textual descriptions that elaborate on the fine-grained category label $c$ (\textit{e.g.}, \textit{switch leap with one turn}), enhanced by a text extender $\mathcal{E}$ (GPT-4~\cite{achiam2023gpt} here) to make them more comprehensible to the model \cite{wei2023cat};
During inference, the Gaussian noise is fed into the trained framework $\mathcal{F}$.
The output is fine-grained action videos $ V^{out} = \{\Tilde{f}_i\}_{i=1}^{T}$, conditioned on the textual and 2D skeletal guidance, denoted as $V^{out} = \mathcal{F}(Noise, D, S^{2D})$.


\vspace{-1.3em}
\subsubsection*{\ding{113} Overall Pipeline.} The whole pipeline is illustrated in Fig.~\ref{fig:2}.
Given video frames, FinePhys first performs online 2D pose estimation, producing the 2D skeleton sequence $S^{2D} \in \mathbb{R}^{T \times J \times 2}$. This sequence serves as an additional modality, providing a compact and bio-structured representation of human actions. Subsequently, these 2D poses undergo an in-context learning process for dimensional lifting, resulting in a data-driven 3D skeleton sequence $S^{3D}_{dd} \in \mathbb{R}^{T \times J \times 3}$.
However, most online estimators struggle to accurately estimate 2D poses, especially for complex movements like gymnastics~\cite{shao2020finegym, huang2025sefar}, leading to noisy 2D inputs. Moreover, the data-driven 2D-to-3D transformation lacks physical interpretability, making the estimation unreliable.
To address these issues, we design a physics-based module called PhysNet, which \textit{re-estimates} the 3D motion dynamics by calculating bidirectional second-order temporal variations (\textit{i.e.}, accelerations) using well-established Euler-Lagrange equations, whose parameters are explicitly predicted.
The refined 3D skeleton sequences, termed physically predicted 3D skeletons and denoted by $S^{3D}_{pp} \in \mathbb{R}^{T \times J \times 3}$, are fused with $S^{3D}_{dd}$ and then projected back to 2D to produce multi-scale latent maps. These skeletal heatmaps are integrated into various stages of the 3D-UNet architecture to guide the video generation process.

For efficient tuning, we incorporate LoRA modules~\cite{guo2023animatediff, zhao2025motiondirector} into the 3D-UNet structure. Specifically, for a weight matrix $W \in \mathbb{R}^{d \times k}$, LoRA employs a low-rank factorization technique to update the original parameters $W_0 \in \mathbb{R}^{d \times k}$ with two trainable low-rank matrices $A \in \mathbb{R}^{d \times r}$ and $B \in \mathbb{R}^{k \times r}$, where $r$ is a smaller rank, \textit{i.e.}, $W = W_0 + AB^T.$
These lightweight modules effectively utilize the skeletal heatmaps to guide the denoising process.

\subsection{FinePhys: Incorporating Physics}
\vspace{-0.2em}
\label{FinePhys}

The core of FinePhys lies in effectively leveraging the physical principles governing human motions. In the following paragraphs, we elaborate on the strategies and design to incorporate useful physics information into FinePhys:

\vspace{-1.3em}
\subsubsection*{Observational Bias: 2D-to-3D Lifting.}
First, we clarify that the term \textit{bias} here is not negative. Instead, it signifies the physical priors and experiences implicitly encoded in large-scale datasets~\cite{banerjee2023physics}. To transform the 2D layout of human joints into 3D geometry, we employ an in-context learning (ICL) process that leverages these biases.
Since only 2D skeletons are provided by the FineGym dataset~\cite{shao2020finegym}, we first obtain a pseudo 3D prior $\overline{P^{3D}} \in\mathbb{R}^{T\times J\times 3}$, which is the calculated mean 3D skeleton sequence from widely used skeleton datasets, including Human3.6M~\cite{ionescu2013human3} and AMASS~\cite{mahmood2019amass}:
{{\setlength\abovedisplayskip{3pt}
\setlength\belowdisplayskip{3pt}
\begin{align}
    \overline{P^{3D}} = \frac{1}{N} \sum_{i=1}^{N} P^{3D}_i,
\end{align}}

\noindent where $N$ is the total number of samples from the above datasets. 
The ICL module requires a few demonstration examples, typically input-output pairs, to form the prompts. In our approach, the prompts $\mathcal{P} = \{ P^{2D}, P^{3D} \}$ consist of ground-truth 2D-3D skeleton pairs randomly selected from Human3.6M. 
The query $\mathcal{Q} = \{ S^{2D}, \overline{P^{3D}} \}$ is composed of detected 2D poses from a FineGym video paired with the previously obtained 3D prior.
The ICL process would generate data-driven 3D skeleton sequences, denoted by:
{{\setlength\abovedisplayskip{3pt}
\setlength\belowdisplayskip{2pt}
\begin{align}
    S^{3D}_{dd} = {\mathbf{Trans}}^{{2D \rightarrow 3D}} \; (\mathcal{P}, \mathcal{Q}),
\end{align}}

\noindent where ${\mathbf{Trans}}^{2D \rightarrow 3D}$ is a two-stream transformer composed of several spatial and temporal blocks as in~\cite{zhu2023motionbert}. 
This process is illustrated in the lower left of Fig.~\ref{fig:2}.
Note that the whole 2D-to-3D lifting procedure benefits directly from the observed data, and the trainable module is expected to capture the underlying physical structures and rules, such as the relationship of limbs, the anatomical limits of joints, the spatial layout of the 3D human body, etc.
\vspace{-2.5em}
\subsubsection*{Inductive Bias: Physics-Informed Module.}
Recent study \cite{kadkhodaiegeneralization} demonstrates that high-quality generative results are achieved through strong inductive biases incorporated within meticulously designed neural networks. Accordingly, we integrate a PhysNet module into our framework to effectively exploit Lagrangian mechanics, as illustrated in Fig.~\ref{fig:3}.
Our primary objective is to estimate the physical terms in the Euler-Lagrange equations to compute temporal variations. To accomplish this, given the data-driven 3D skeleton data $S^{3D}_{dd}$, 
we embed both global and local temporal dynamics using distinct encoders (\textit{i.e.}, global head $\mathbb{E}_\theta^{(g)}$ and local head $\mathbb{E}_\theta^{(l)}$):
\setlength\abovedisplayskip{3pt}
\setlength\belowdisplayskip{3pt}
\begin{align}
    q^{(g)}_t &= \mathbb{E}_\theta^{(g)} (\, \{ S^{3D}_{dd}(t) \}_{t=1}^{T} \,)\quad \in\mathbb{R}^{T\times (J\times 3)}, \\
    q^{(l)(\rightarrow)}_t &= \mathbb{E}_\theta^{(l)} (\, \{ S^{3D}_{dd}(t-2), S^{3D}_{dd}(t-1), S^{3D}_{dd}(t) \} \, ), \\
    q^{(l)(\leftarrow)}_t &= \mathbb{E}_\theta^{(l)} (\, \{ S^{3D}_{dd}(t), S^{3D}_{dd}(t-1), S^{3D}_{dd}(t-2) \} \, ),
\end{align}
where  $ q^{(l)(\rightarrow)}_t$ and $ q^{(l)(\leftarrow)}_t$ represent the forward and reverse temporal direction, respectively.
All subsequent computations are performed bidirectionally, with forward updates starting from the first three frames and reverse updates from the last three frames. 
For simplicity, the temporal direction ($\rightarrow$ and $\leftarrow$) is omitted in the notation.
Then we obtain the temporal state vector $q_t$ at each time step $t$ by fusing $q^{global}_t$ and $q^{local}_t$.
Using  $q_t$ as input, we estimate the corresponding parameters within the Euler-Lagrange equation:
\begin{align}
    M(q_t) \ddot{q_t} = J(q_t, \dot{q_t})- C (q_t, \dot{q_t}).
\end{align}}

\noindent Specifically, for the generalized forces $\hat{J}_t$, and the joint constraints $\hat{C}_t$ in forward updating:
\begin{align}
    \hat{J}_{t} = \mathbb{E}_{J}^{\text{Phys}}(q_t) \quad \in\mathbb{R}^{51}, \\
    \hat{C}_{t} = \mathbb{E}_{C}^{\text{Phys}}(q_t) \quad \in\mathbb{R}^{51}.
\end{align}}

\noindent Estimating the inverse inertia matrix $M^{-1} \in \mathbb{R}^{51 \times 51}$ poses significant challenges due to its high dimensionality. As discussed in~\cite{kadkhodaiegeneralization}, imposing strong constraints or priors on the hypothesis space facilitates this estimation. 
Thus, we estimate $M^{-1}$ in two steps: (1) assume symmetry, and (2) incorporate Gaussian noise. The symmetry assumption is intuitively based on the prior knowledge that inertia tensors and mass matrices are typically symmetric in structural systems. Consequently, we define:
{{\setlength\abovedisplayskip{3pt}
\setlength\belowdisplayskip{3pt}
\begin{align}
    {(\hat{M}_t^{-1})}^{\Delta} = \mathbb{E}_{M}^{\text{Phys}}(q_t) \quad & \in\mathbb{R}^{51 \times 26}, \\
    \hat{M}^{-1}_t = \mathcal{S} (\: {(\hat{M}_t^{-1})}^{\Delta} \:) \quad & \in\mathbb{R}^{51 \times 51},
\end{align}}

\noindent where ${(\hat{M}_{t}^{-1})}^{\Delta} $ denotes the upper triangular matrix of $(\hat{M}^{-1}_{t})$, and $\mathcal{S}$ is the symmetric operation.

However, motion can disrupt the body's symmetry. To account for this, we introduce a noise parameter $\hat{N}_t$:
{{\setlength\abovedisplayskip{3pt}
\setlength\belowdisplayskip{3pt}
\begin{align}
    \hat{N}_t^{[i]} &=  \mathcal{G} (\, \mathbb{E}^{\text{Phys}}_N(q_t) \,), \in\mathbb{R}^{51}, \\
    \hat{N}_t &= \{\hat{N}_t^{[i]} \} \in\mathbb{R}^{51 \times 51},  i = \{1,...,51\}, 
\end{align}}} 

\noindent where $\mathcal{G}$ is a Gaussian sampling process with variance $\sigma^2 = 1$, adding small random noise to each column vector $\hat{N}_t^{[i]}$ in $\hat{N}_t$.
With the estimated parameters, we compute:
{{\setlength\abovedisplayskip{3pt}
\setlength\belowdisplayskip{1pt}
\begin{align}
\ddot{q}_{t} &= ((\hat{M}^{-1}_t) + \hat{N}_t)\cdot (\hat{J}_t - \hat{C}_t), 
\end{align}}
}

\noindent and then obtain the future states with the second-order central difference formula
 $\ddot{q}_t\approx \frac{q_{t+1} - 2q_{t} + q_{t-1}}{(dt)^2}$:
  {{\setlength\abovedisplayskip{3pt}
\setlength\belowdisplayskip{3pt}
 \begin{align}
    \hat{q}_{t+1} = \ddot{q}_{t} \cdot (dt)^2 + 2q_{t} - q_{t-1} \in\mathbb{R}^{T\times 51}.
\end{align}}}

\noindent Since updates occur bidirectionally, each middle timestep (excluding the first and last three) yields two estimates: $\hat{q}_{t+1}^{\rightarrow}$ from forward updating and $\hat{q}_{t+1}^{\leftarrow}$  from reverse updating. We average these estimates:
$\hat{q}_{t+1} = (\hat{q}_{t+1}^{\rightarrow} + \hat{q}_{t+1}^{\leftarrow}) / 2 $.
This results of state sequence $\hat{q} = \{ \hat{q}_{t} \}_{t=1}^T$ is input to the pose decoder $\mathbb{E}^{\text{pose}}$ for the physically predicted 3D skeletons:
\begin{align}
    S^{3D}_{pp} = \mathbb{E}^{\text{pose}} (\hat{q})\quad \in\mathbb{R}^{T\times 17\times 3}.
\end{align}

\begin{figure}
    \centering
    \includegraphics[width=1.0\linewidth]{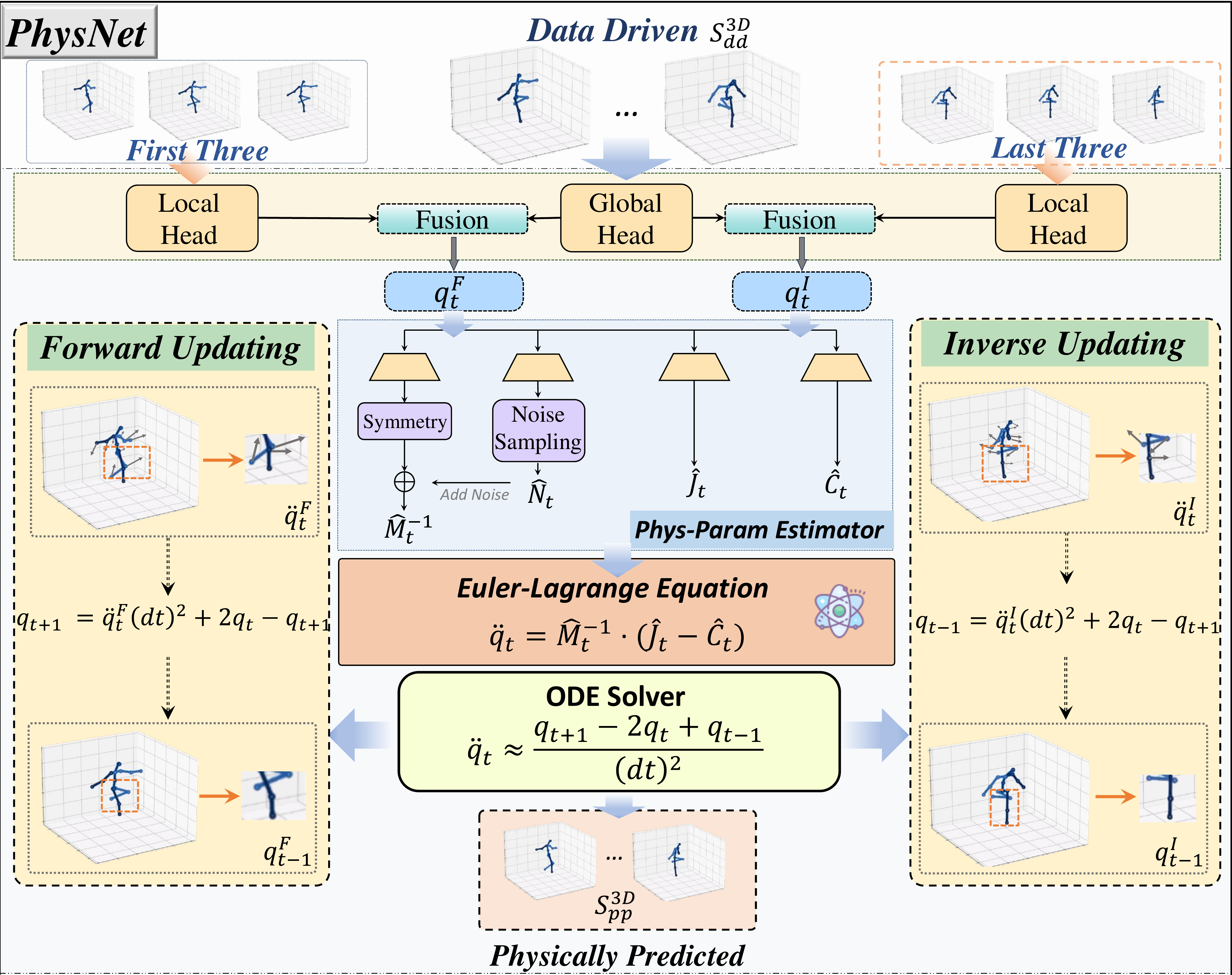}
    \vspace{-1.8em}
    \caption{\textbf{The PhysNet Module.}
    Given the input $S_{dd}^{3D}$, PhysNet leverages both global and local temporal dynamics in a bi-directional manner to estimate the terms of the Euler-Lagrange equations. By integrating with an ODE solver, the module can predict future and past states, thereby enhancing the original $S_{dd}^{3D}$ across both temporal directions and producing physically predicted 3D sequences, denoted as $S_{pp}^{3D}$.}
    \label{fig:3}
    \vspace{-1.2em}
\end{figure}

\vspace{-1em}
\subsubsection*{Learning Bias: Optimization Objectives.}
\textit{``Learning bias"} refers to the incorporation of prior physical knowledge through penalty constraints. 
Specifically, the training of FinePhys involves the following stages, each
employing loss functions adhering to the underlying physics:

\noindent\ding{172} \textbf{The pre-training stage}
aims to enhance the accuracy of 3D pose estimation from 2D inputs. 
To achieve this, we utilize large-scale datasets that provide ground truth 3D poses $S^{3D}$, such as Human3.6M and AMASS. 
The 2D poses are processed by both the in-context learning module and the PhysNet module, and are subsequently fused to obtain the estimated 3D poses: $\hat{S}^{3D} = \mathcal{F}(S^{3D}_{dd}, S^{3D}_{pp})$.
Additionally, within the PhysNet module, we introduce an auxiliary loss, $\mathcal{L}_{\text{noise}} = \sum_{t=1}^{T} \|\hat{N}_t\|_F$, to constrain the noise vector $\hat{N}_t$ that is added to perturb the symmetry of $\hat{M}^{-1}_t$ ($t$ denotes time steps).
The loss during this stage is calculated as:
{{\setlength\abovedisplayskip{3pt}
\setlength\belowdisplayskip{3pt}
\begin{align}
    \mathcal{L}_{3D} = \sum_{t=1}^T \sum_{j=1}^J \|\hat{S}^{3D}_{t,j} - S^{3D}_{t,j}\|_2^2 + \mathcal{L}_{\text{noise}}, 
\end{align}}
\noindent where $j$ represents human joints, with $J=17$.

\noindent\ding{173} \textbf{The fine-tuning stage} utilizes fine-grained human action videos from FineGym~\cite{shao2020finegym}, which present greater challenges due to rapid temporal dynamics and significant body deformations. Additionally, FineGym does not provide ground truth 3D poses. Therefore, we project the estimated 3D poses $\hat{S}^{3D}_{t,j}$ into 2D using the projection module $\mathcal{P}$ and compute the re-projection loss in the 2D space:
\setlength\abovedisplayskip{3pt}
\setlength\belowdisplayskip{3pt}
\begin{equation}
\begin{aligned}
    \mathcal{L}_{2D} = \sum_{t=1}^T \sum_{j=1}^J \|\mathcal{P}(\hat{S}^{3D}_{t,j}) - {S}^{2D}_{t,j}\|_2^2 + \mathcal{L}_{\text{noise}},
\end{aligned}
\end{equation}
and $\mathcal{L}_{\text{noise}}$ is as defined above.

\vspace{0.1em}
\noindent\ding{174} Finally, \textbf{the generation stage} involves generating video frames, wherein the entire framework is trained end-to-end using generation losses.
These losses include the spatial loss $\mathcal{L}_{\text{spat}}$, the temporal loss $\mathcal{L}_{\text{temp}}$, and the appearance-debiased temporal loss $\mathcal{L}_{\text{ad-temp}}$, resulting in:
\begin{align}
    &\mathcal{L}_{\text{spat}} = \mathbb{E}_{z_0, c, \epsilon, t, i\sim\mathcal{U}(1, T)}[\|\epsilon - \epsilon_\theta(z_{t, i}, t, \tau_\theta(c))\|_2^2], \\
    &\mathcal{L}_{\text{temp}} = \mathbb{E}_{z_0, c, \epsilon, t}[\|\epsilon - \epsilon_\theta(z_{t}, t, \tau_\theta(c))\|_2^2], \\
    &\mathcal{L}_{\text{ad-temp}} = \mathbb{E}_{z_0, c, \epsilon, t}[\|\phi(\epsilon) - \phi(\epsilon_\theta(z_{t}, t, \tau_\theta(c)))\|_2^2], \\
    &\mathcal{L}_{\text{video}} = \mathcal{L}_{\text{spat}} + \mathcal{L}_{\text{temp}} + \mathcal{L}_{\text{ad-temp}}, 
\end{align}
\noindent where $z_{t, i}$ denotes the $i_{th}$ frame of $z_t$, and $\phi$ is the debiasing operator as described in~\cite{zhao2025motiondirector}.

\section{Experiments}
\subsection{Experimental Setup}

\begingroup
\setlength{\tabcolsep}{5pt}
\begin{table*}
\renewcommand{\arraystretch}{1}
    \centering
    \caption{\textbf{Comparison with state-of-the-art methods on two subsets of FineGym (FX-JUMP and FX-TURN).} In this table, ``T" within \textbf{Input} denotes textual prompt, while ``P", ``I", ``D", and ``C" represent the pose, initial frame, depth map, and canny, respectively.
    In all cases, our FinePhys outperforms various baselines based on diverse conditions by a large margin in terms of more reliable metrics, CLIP-SIM*, 
    and User Study (the insufficiency of the CLIP-SIM has been shown).
    }
    \vspace{-0.8em}
    \centering
    \scriptsize
     \begin{tabular}{lc|ccc|cc|cc|ccc}
       \hlineB{2.5}
       \multirow{2}{*}{\textbf{Method}} & \multirow{2}{*}{\textbf{Input}} & \multicolumn{3}{|c|}{\textbf{User Study}} & \multicolumn{2}{c|}{\textbf{CLIP-SIM*}} & \multirow{2}{*}{\textbf{PickScore$\uparrow$}}& \multirow{2}{*}{\textbf{FVD$\downarrow$}} & \multicolumn{3}{c}{\textbf{CLIP-SIM}} \\
       \cline{3-5}\cline{6-7}\cline{10-12}
        &  & \multicolumn{1}{|c}{\textbf{Text.$\uparrow$}} & \textbf{Domain.$\uparrow$} & \multicolumn{1}{c|}{\textbf{Smooth.$\uparrow$}} & \textbf{Domain.$\uparrow$} & \textbf{Smooth.$\uparrow$} & & &\textbf{Text.$\uparrow$} & \textbf{Domain.$\uparrow$} & \textbf{Smooth.$\uparrow$} \\
       \hlineB{2}
       \rowcolor{gray!20} \multicolumn{12}{c}{{\textit{w/o finetuning on FineGym}}} \\
       Control-A-Video \cite{chen2023control} \textit{\textcolor{gray}{{\fontsize{6}{12}\selectfont arXiv'23}}} & T+D & 3.43 & 3.13 & 3.37 & 0.697 & 0.706 & 18.995 & 632.68 & \textcolor{gray}{26.456} & \textcolor{gray}{0.640}& \textcolor{gray}{0.900} \\
       Control-A-Video \cite{chen2023control} \textit{\textcolor{gray}{{\fontsize{6}{12}\selectfont arXiv'23}}} & T+C & 3.10 & 2.63 & 2.60 & 0.508 & 0.520 & 18.339 & 637.79 & \textcolor{gray}{18.755} & \textcolor{gray}{0.591} & \textcolor{gray}{0.899} \\
       VideoCrafter1 \cite{chen2023videocrafter1} \textit{\textcolor{gray}{{\fontsize{6}{12}\selectfont arXiv'23}}}& T+I & 2.53 & 2.57 & 2.60 & 0.685 & 0.682 & 18.750 & 510.09 & \textcolor{gray}{24.821} & \textcolor{gray}{0.591} & \textcolor{gray}{0.869} \\
       Text2Video-Zero~\cite{khachatryan2023text2video} \textit{\textcolor{gray}{{\fontsize{6}{12}\selectfont ICCV'23}}} & T & 1.93 & 1.80 & 2.00 & 0.501 & 0.509 & 17.827 & 897.61 & \textcolor{gray}{19.368} & \textcolor{gray}{0.613} & \textcolor{gray}{0.921}  \\
       Text2Video-Zero~\cite{khachatryan2023text2video} \textit{\textcolor{gray}{{\fontsize{6}{12}\selectfont ICCV'23}}} & T+P & 1.83 & 1.90 & 1.67 & 0.481 & 0.484 & 17.659 & 904.50 & \textcolor{gray}{16.725} & \textcolor{gray}{0.620} & \textcolor{gray}{0.978} \\
       Latte \cite{ma2024latte} \textit{\textcolor{gray}{{\fontsize{6}{12}\selectfont arXiv'24}}}& T & 1.97 & 2.03 & 2.13 & 0.681 & 0.675 & 19.421 & 590.41 & \textcolor{gray}{27.197} & \textcolor{gray}{0.693} & \textcolor{gray}{0.906} \\
       Follow-Your-Pose \cite{ma2024follow} \textit{\textcolor{gray}{{\fontsize{6}{12}\selectfont AAAI'24}}}& T+P & 2.20 & 2.13 & 2.37 &  0.612 & 0.627 & 18.680 & 640.12 & \textcolor{gray}{27.198} & \textcolor{gray}{0.647} & \textcolor{gray}{0.888} \\
       AnimateDiff \cite{guo2023animatediff} \textit{\textcolor{gray}{{\fontsize{6}{12}\selectfont ICLR'24}}}& T & 2.20 & 2.57 & 2.33 &  0.686 & 0.686 & 19.468 & 704.74 & \textcolor{gray}{28.629} & \textcolor{gray}{0.669} & \textcolor{gray}{0.938} \\
       AnimateDiff \cite{guo2023animatediff} \textit{\textcolor{gray}{{\fontsize{6}{12}\selectfont ICLR'24}}}& T+I & 2.73 & 2.60 & 2.93 & 0.684 & 0.699 & 19.362 & 535.79 & \textcolor{gray}{26.604} & \textcolor{gray}{0.629} & \textcolor{gray}{0.881} \\
       VideoCrafter2 \cite{chen2024videocrafter2} \textit{\textcolor{gray}{{\fontsize{6}{12}\selectfont CVPR'24}}}& T & 2.23 & 2.50 & 2.60 & 0.660 & 0.651 & \textbf{20.023} & 697.73 & \textcolor{gray}{26.296} & \textcolor{gray}{0.714} & \textcolor{gray}{0.964}  \\
       \hlineB{2}
       \rowcolor{gray!20} \multicolumn{12}{c}{{\textit{w/ finetuning on FineGym}}} \\
       Follow-Your-Pose \cite{ma2024follow} \textit{\textcolor{gray}{{\fontsize{6}{12}\selectfont AAAI'24}}} & T+P  & 2.67 & 2.53 & 2.57 & 0.709 & 0.727 & 19.360 & 506.26 & \textcolor{gray}{28.929} & \textcolor{gray}{0.587} & \textcolor{gray}{0.905} \\
       AnimateDiff \cite{guo2023animatediff} \textit{\textcolor{gray}{{\fontsize{6}{12}\selectfont ICLR'24}}}& T  & 3.17 & 3.07 & 2.97 & 0.728 & 0.752 & 19.070 & 522.14 & \textcolor{gray}{26.791} & \textcolor{gray}{0.546} & \textcolor{gray}{0.880} \\
       AnimateDiff \cite{guo2023animatediff} \textit{\textcolor{gray}{{\fontsize{6}{12}\selectfont ICLR'24}}}& T+I & 3.20 & 3.20 & 3.17 & 0.769 & 0.793 & 19.705 & 529.38 & \textcolor{gray}{27.033} & \textcolor{gray}{0.583} & \textcolor{gray}{0.873} \\
       \hline
       \rowcolor{cyan!10}
       FinePhys (Ours)  & T+P &  \textbf{4.13} & \textbf{3.86} & \textbf{4.03} & \textbf{0.826} & \textbf{0.833} & \underline{19.941} & \textbf{484.49} & \textcolor{gray}{27.073} & \textcolor{gray}{0.520} & \textcolor{gray}{0.939} \\
       \hlineB{2.5}
    \end{tabular}
    \label{tab:1}
\end{table*}
\endgroup

\vspace{-0.2em}
\subsubsection*{Training and Datasets.}
\ding{182} Pre-training Datasets: We first train the skeletal heatmap encoder on the HumanArt~\cite{ju2023humansd} dataset, using skeletons and images as inputs. 
This dataset contains a large number of human skeleton-image pairs. 
To train our 
2D-to-3D module, we collect diverse and realistic 3D human motion data for the pretraining phase of skeleton modeling, following the design of~\cite{wang2024skeleton}, including Human3.6M~\cite{ionescu2013human3} and AMASS~\cite{von2018recovering}.
\ding{183} Fine-grained Action Datasets: we construct three subsets from FineGym~\cite{shao2020finegym}: FX-JUMP, FX-TURN, and FX-SALTO, derived from the Floor Exercise event in FineGym. These subsets possess different motion characteristics, 
and are used for tuning the FinePhys framework as well as for validation.
Further details are provided in the Supplementary.

\begin{figure}[t]
    \centering
    \includegraphics[width=1.0\linewidth]{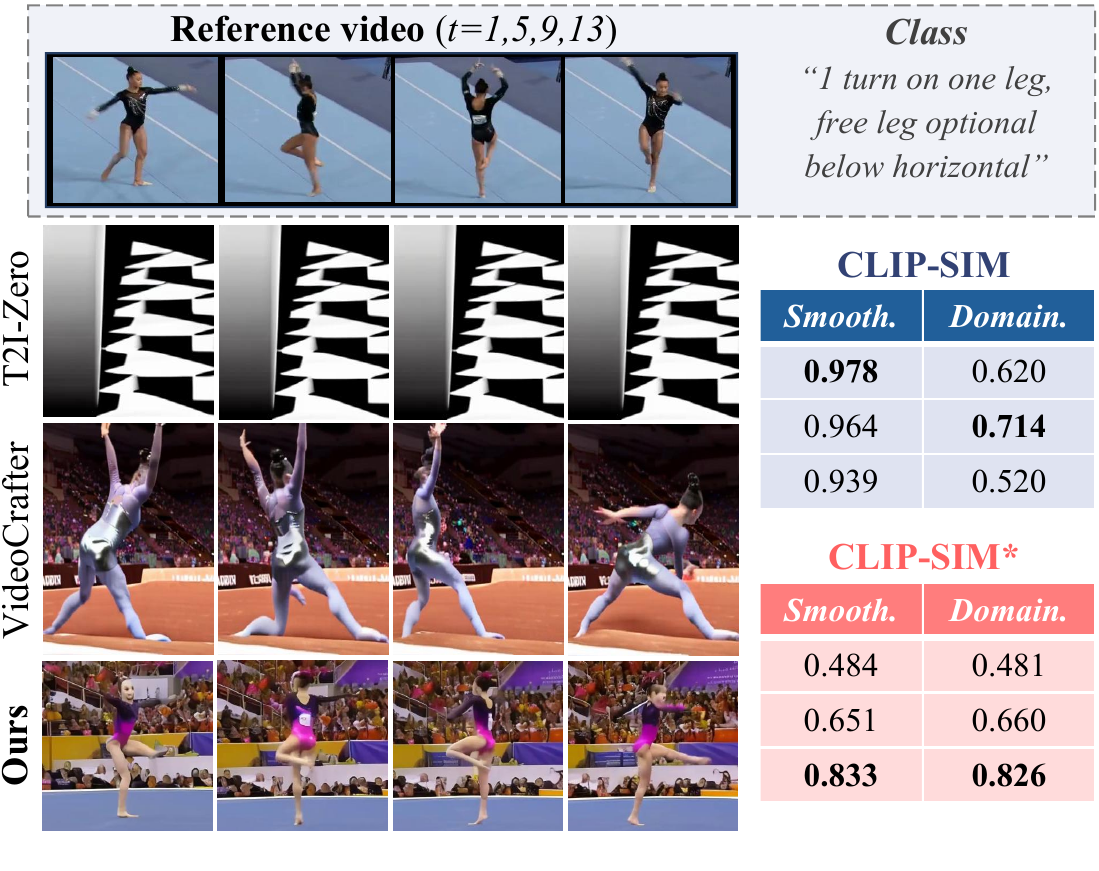}
    \vspace{-2.6em}
    \caption{
    \textbf{Original CLIP-SIM} metrics \textit{fail} to evaluate the generated results (\textit{e.g.}, 
    T2I-Zero produces entirely irrelevant outputs yet achieves the highest smooth score according to the original CLIP-SIM. 
    In contrast, our \textbf{enhanced CLIP-SIM*} provides a more reliable evaluation that better aligns with human judgment.
    }
    \label{fig:metric}
    \vspace{-1em}
\end{figure}

\vspace{-1.2em}
\subsubsection*{Implementation Details.}
For the generation backbone, 
we use the official codebase of Stable Diffusion v1.5~\cite{rombach2022high} and the motion module checkpoints from AnimateDiff~\cite{guo2023animatediff}.
We extract video clips at a resolution of $384\times 384$ pixels, consisting of $16$ frames each for training. 
Using FineGym, we selected $35$ classes from  FX-JUMP, FX-TURN, and FX-SALTO, totaling $350$ videos.
Specifically,
We first train our skeletal heatmap encoder for $54$k steps on the real-human part of HumanArt~\cite{ju2023humansd}, following Follow-Your-Pose~\cite{ma2024follow}.
The 2D-to-3D module together with PhysNet module is pre-trained on Human36M and AMASS (with 3D pose annotations) for 10 epochs.
Then the PhysNet module and 2D projection module are fine-tuned based on 2D skeletons detected from FineGym online.
We also tune the LoRA module for $8$k steps on the fine-grained datasets FX-JUMP, FX-TURN, and FX-SALTO.

\vspace{-0.2em}
\subsection{Main Results}
\vspace{-0.2em}
The evaluations were conducted on three fine-grained human action subsets drawn from FineGym: FX-JUMP, FX-TURN, and FX-SALTO. These subsets include challenging gymnastics actions executed by professional gymnasts.
Prior to presenting detailed method comparisons, we introduce the evaluation metrics used for quantitative assessment and discuss specific anomalies observed when using them to evaluate fine-grained human action video generation.

\vspace{-1.4em}
\subsubsection*{Evaluation Metrics.}
\ding{182} \textit{Automatic Metrics}: We use PickScore \cite{kirstain2023pick} to measure the alignment between video frames and text prompts, CLIP Domain and CLIP Smooth Similarity \cite{radford2021learning} to evaluate semantic similarity and embedding stability, and Fréchet Video Distance (FVD) \cite{unterthiner2018towards} for video quality assessment.
\ding{183} \textit{User Study:} 
We conducted a user study, leveraging human sensitivity and accuracy in assessing motion plausibility, bio-structure preservation, and visual acceptability. Specifically, participants were presented with a set of videos simultaneously, including one video generated by our method alongside those from baseline methods. For each video, they rated the consistency of the following aspects on a scale from $1$ to $5$:
\ding{172} \text{Text Alignment}, 
\ding{173} \text{Domain Consistency}, and 
\ding{174} \text{Smooth Stability}.
The mean opinion score (MOS) is reported as the final result.

\vspace{-1.4em}
\subsubsection*{Discussion and Improved Metrics.}
Existing metrics may be unreliable for evaluating video generation results~\cite{kwon2024harivo}.
In our experiments, we found that these evaluation anomalies are even more pronounced for fine-grained human action video generation. Below, we first elaborate limitations of the original CLIP-SIM and then introduce an improved version for more accurate evaluation.
(1) The original CLIP-SIM metric measures semantic consistency (SC), domain consistency (DC), and temporal consistency (TC) are achieved by calculating similarities between text-to-video, image-to-video, and video-to-video, respectively. 
However, for \textbf{TC}, fine-grained semantics are not well captured by CLIP~\cite{tong2024eyes}, making it less effective.
Additionally, \textbf{DC} relies on reference images generated by Stable Diffusion, which may yield high scores for entirely irrelevant visual content, as shown in Fig.~\ref{fig:metric}. 
Moreover, the original \textbf{TC} only considers inter-frame similarity and completely ignores changing motion dynamics. 
Therefore, the original CLIP-SIM metrics are inadequate for evaluating fine-grained human action video generation, as depicted in Fig.~\ref{fig:metric}.
(2) To provide a more reliable evaluation, we introduce an improved version of the CLIP metrics:
{{\setlength\abovedisplayskip{3pt}
\setlength\belowdisplayskip{3pt}
\begin{align}
    \text{CLIP}_{\text{DS}}^* (\Tilde{V}, \{I_j\}) = \frac{1}{N}\frac{1}{T}\sum_{t=1}^T \sum_{j=1}^N \text{CLIP}(\hat{V}(t), I_j), \\
    \text{CLIP}_{\text{TC}}^*(\Tilde{V},V^{\text{Ref}})=\sum_{k=1}^K\sum_{l=1}^M \text{CLIP}(\Tilde{V}(k), V^{\text{Ref}}_l(k)),
\end{align}}

\noindent where $\Tilde{V}$ are generated videos and $\{I_j\}_{j=1}^N$ are $N$ frames sampled from FineGym actions. 
For $\text{CLIP}_{\text{TC}}^*$, $V^{\text{Ref}}$ consists of randomly chosen reference videos from FineGym, and the calculation employs a multi-step sampling strategy. As illustrated in Fig.~\ref{fig:metric}, the new metrics provide a more accurate evaluation of the generated results. Further details on these metrics are provided in the Supplementary Material.

\begin{figure}[t]
    \centering
    \includegraphics[width=1.0\linewidth]{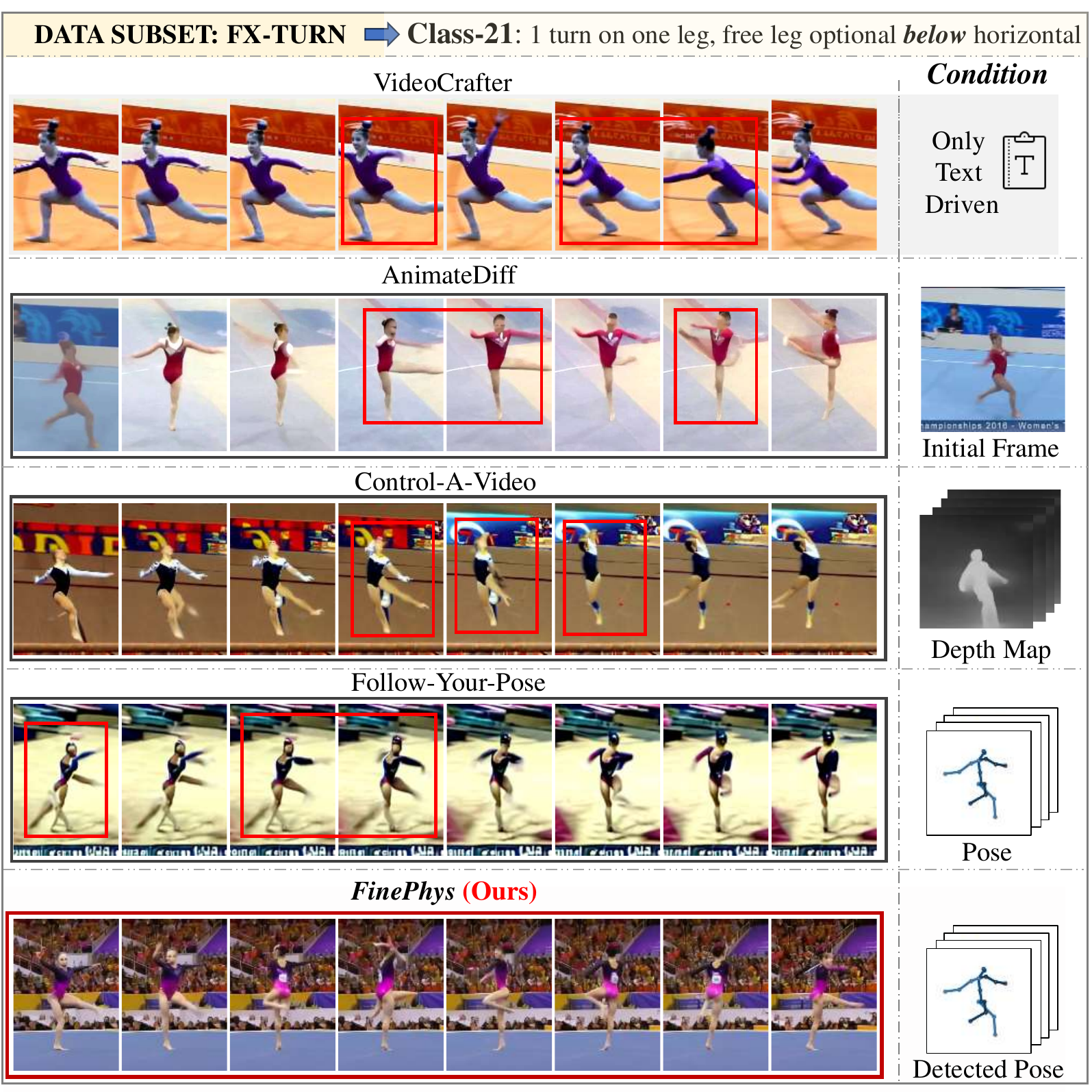}
    \vspace{-1.9em}
    \caption{\textbf{Qualitative Results.}
    Compared to other baselines, FinePhys demonstrates superior performance in understanding complex, fine-grained semantics, maintaining biomechanical consistency, and adhering to physical principles.
    }
    \label{fig:qualitative}
    \vspace{-0.6em}
\end{figure}

\vspace{-1.4em}
\subsubsection*{Quantitative Comparison with Baselines.}
Using the aforementioned metrics, we evaluate FinePhys against competitive baselines, including Control-A-Video~\cite{chen2023control}, VideoCrafter1/2~\cite{chen2023videocrafter1, chen2024videocrafter2}, Text2Video-Zero~\cite{khachatryan2023text2video}, Latte~\cite{ma2024latte}, Follow-Your-Pose~\cite{ma2024follow}, and AnimateDiff~\cite{guo2023animatediff}, on generating fine-grained human actions. 
Results are presented in Tab.~\ref{tab:1}. 
Despite the insufficiency of the original CLIP-SIM
FinePhys significantly outperforms these baselines on the improved CLIP-SIM* metrics and the user study (which are widely recognized as more credible metrics for evaluating video generation results),
demonstrating its superior ability to understand fine-grained human actions and generate more physically plausible motions. 

\vspace{-1.4em}
\subsubsection*{Qualitative Analysis.} 
We visualize the generated results of FinePhys alongside baseline methods, including those with additional conditions~\cite{guo2023animatediff,ma2024follow,chen2023control} and purely text-driven approaches~\cite{chen2023videocrafter1}, as shown in Fig.~\ref{fig:instance} and Fig.~\ref{fig:qualitative}. 
We observe that these strong baselines struggle to generate satisfactory results:
VideoCrafter2~\cite{chen2024videocrafter2} (only textual conditions) frequently exhibits dramatic flaws like no actions and character anomalies;
AnimateDiff~\cite{guo2023animatediff} shows neglectable inconsistencies such as character changes and incorrect semantics for \textit{``below"};
Control-A-Video~\cite{chen2023control} (conditioned on depth maps) fails to interpret the action dynamics correctly;
Follow-Your-Pose~\cite{ma2024follow} also relies on 2D skeletons, displays limb distortions and low visual quality.
In contrast, FinePhys effectively interprets complex motion dynamics (\textit{e.g.}, jumps with 0.5 turns), understands fine-grained semantics (\textit{e.g.}, “turning in stand position with leg below horizontal”), and better preserves the bio-structure of human body, being able to generate physically plausible actions.



\noindent\begingroup
\setlength{\tabcolsep}{1.1pt}
\begin{table}[t]
  \vspace{-0.4em}
  \centering
  \caption{\textbf{Quantitative evaluation of different pose results.} \textit{Left Part} shows results on Human3.6M in 3D spaces. \textit{Right Part} shows results on FineGym in 2D spaces.
  Metrics include mean per joint position error (MPJPE), Normalized MPJPE (N-MPJPE), and mean per-joint velocity error (MPJVE). Note that $S_{dd}^{3D}$ and $S_{pp}^{3D}$ will be projected back into 2D for evaluation on FineGym.}
  \vspace{-1em}
  \scriptsize
  \renewcommand{\arraystretch}{1.3}
  \begin{tabular}{l|>{\centering\arraybackslash}p{1cm}>{\centering\arraybackslash}p{1.23cm}>{\centering\arraybackslash}p{1cm}|>{\centering\arraybackslash}p{1cm}>{\centering\arraybackslash}p{1.23cm}>{\centering\arraybackslash}p{1cm}}
    \hlineB{2.5}
     \multirow{2}{*}{\textbf{Pose Results}} & \multicolumn{3}{c|}{\textbf{3D Eval. on Human3.6M}} & \multicolumn{3}{c}{\textbf{2D Eval. on FineGym}} \\
    \cline{2-7}
      & {\fontsize{6}{12}\selectfont\textbf{MPJPE}$\downarrow$} & {\fontsize{6}{12}\selectfont\textbf{N-MPJPE}$\downarrow$} & {\fontsize{6}{12}\selectfont\textbf{MPVPE}$\downarrow$} & {\fontsize{6}{12}\selectfont\textbf{MPJPE}$\downarrow$} & {\fontsize{6}{12}\selectfont\textbf{N-MPJPE}$\downarrow$} & {\fontsize{6}{12}\selectfont\textbf{MPVPE}$\downarrow$} \\
    \hlineB{2}
    \rowcolor{yellow!10}
    $S^{2D}_{\text{detect}}$ & - & - & -  & 0.918 & 0.254 & 0.379 \\
    \rowcolor{yellow!10}
    $S_{dd}$ & 0.048 & 0.046 & 0.020 & 0.229 & 0.215 & 0.108 \\
    \rowcolor{yellow!10}
    $S_{pp}$ & 0.068 & 0.066 & 0.035 & 0.237 & 0.213 & 0.097 \\
    \rowcolor{yellow!10}
    $S_{dd}$+$S_{\text{MLP}}$ & 0.065 & 0.060 & 0.025 & 0.243 & 0.237 & 0.140 \\
    \rowcolor{cyan!10}
    $S_{dd}$+$S_{pp}$ & \textbf{0.046} &\textbf{0.044} &\textbf{0.018} & \textbf{0.178} &\textbf{0.147} &\textbf{0.094} \\
    \hlineB{2.5}
  \end{tabular}
  \label{tab:abla_2d}
  \vspace{-1.2em}
\end{table}
\endgroup

\vspace{-0.8em}
\subsection{Ablations and Analysis}
\subsubsection*{Transformation of skeleton data.} 

We evaluate the pose results obtained from different modules and procedures by conducting the following two sets of experiments:
\ding{172} \textbf{Evaluation in 3D space} is done on Human3.6M (3D pose annotations provided).
The results are presented in Tab.~\ref{tab:abla_2d}. 
Since the actions in Human3.6M primarily involve daily activities with moderate pose variations, the in-context learning module achieves good results ($S^{3D}_{dd}$). 
Additionally, averaging $S^{3D}_{dd}$ and $S^{3D}_{pp}$ reduces estimation errors, indicating that physically predicted poses can mitigate deviations in data-driven estimates and thereby validate our design.
\ding{173} \textbf{Evaluation in 2D space} is performed on FineGym subsets (without 3D pose annotations).
By projecting $S^{3D}_{dd}$ and $S^{3D}_{pp}$ into 2D, we obtain $S^{2D}_{dd}$ and $S^{2D}_{pp}$, respectively, and compare them with the online estimated results $S^{2D}_{\text{detect}}$. As expected, $S^{2D}_{\text{detect}}$ performs poorly due to extreme body deformations and rapid temporal changes. Notably, $S^{2D}_{pp}$ exhibits higher accuracy, further validating the necessity of the PhysNet module for fine-grained action understanding. Combining $S^{2D}_{dd}$ and $S^{2D}_{pp}$ yields the best results, underscoring the significance of each module within our FinePhys.

\vspace{-1.3em}
\subsubsection*{{Robustness on noisy input during inference.}} 
Online 2D pose estimation for fine-grained action videos often produces highly noisy results, as illustrated in Fig.~\ref{fig:noisy_input}. 
Without additional processing and sophisticated designs, conditioning the generation process directly on such noisy pose inputs leads to poor outcomes, as the results from Follow-Your-Pose~\cite{ma2024follow} shown in Fig.~\ref{fig:noisy_gen}. 
However, by leveraging the dimension lifting and PhysNet modules, 
FinePhys effectively restores distorted and missing poses, as shown in Fig.~\ref{fig:noisy_input}.
These mechanisms collectively enhance the robustness of FinePhys when handling noisy pose inputs.

\vspace{-1.3em}
\subsubsection*{Importance of the PhysNet module.} 
\ding{172} First, without PhysNet, we would not obtain $S^{3D}_{pp}$ and $S^{2D}_{pp}$. 
This absence leads to inferior pose guidance, as shown in Tab.~\ref{tab:abla_2d},
and would logically lead to poorer generation results.
\ding{173} To justify our design, we replaced PhysNet with a simple MLP to embed the data-driven poses further. This substitution resulted in significantly worse performance, as demonstrated in the bottom rows of Tab.~\ref{tab:abla_2d}.

\vspace{-1.3em}
\subsubsection*{Limitation and future work.} 
While FinePhys significantly outperforms previous approaches in generating fine-grained human action videos, it represents only an initial step. 
Generating fine-grained actions such as various \textit{salto}, which involve simultaneous body rotations and rapid turns in the air, still remains highly intractable (thus not calculated in Tab.~\ref{tab:1}).
Additionally, generating detailed frames 
may divert focus from the deeper integration of physical principles, thus we plan to utilize simpler scenarios and further explore the modeling of physics in future work.

\begin{figure}[t]
    \centering
    \vspace{-0.4em}
    \includegraphics[width=1.0\linewidth]{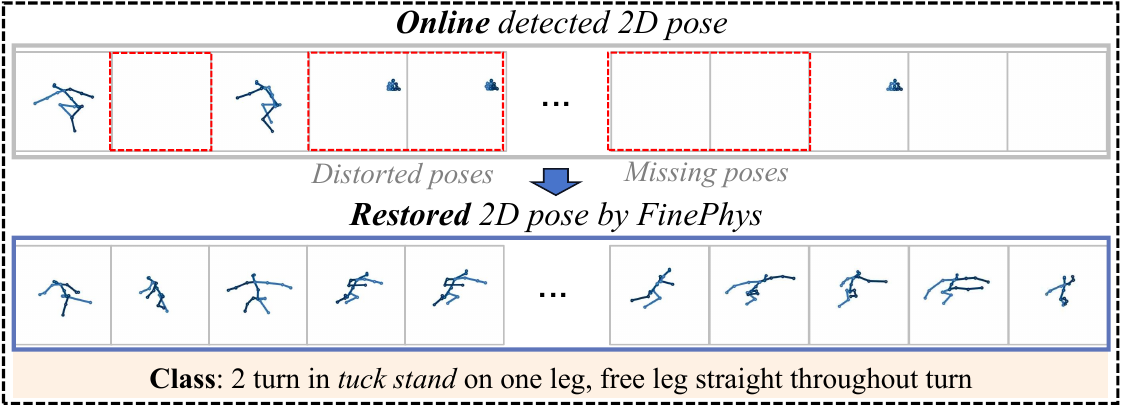}
    \vspace{-1.9em}
    \caption{FinePhys effectively restores distorted and missing poses using the in-context learning and PhysNet modules, thereby providing enhanced skeletal guidance for the generation process.
    }
    \label{fig:noisy_input}
    \vspace{-0.4em}
\end{figure}

\begin{figure}[t]
    \centering
    \includegraphics[width=1.0\linewidth]{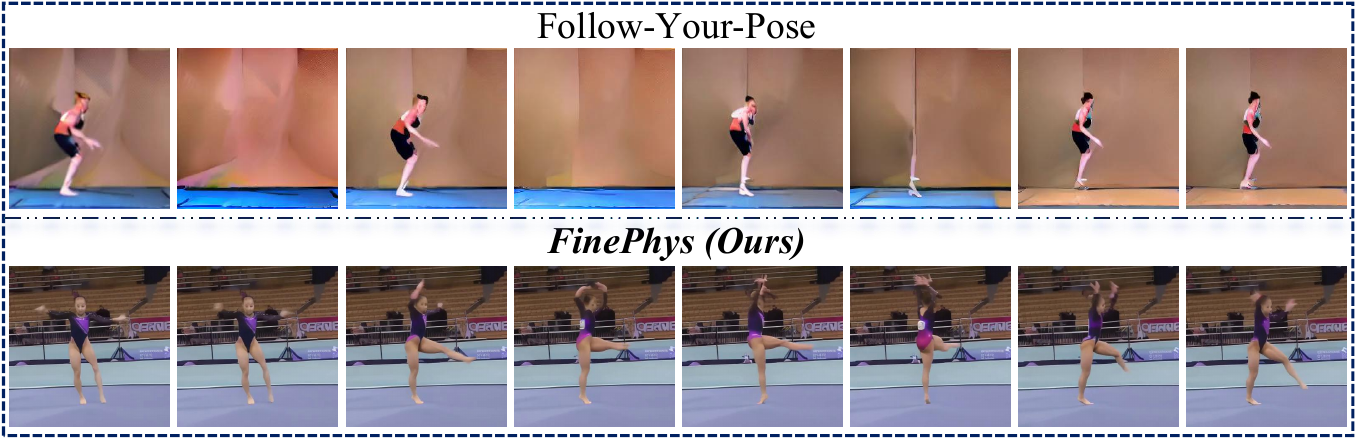}
    \vspace{-1.9em}
    \caption{Generation results from FinePhys and Follow-Your-Pose~\cite{ma2024follow} conditioned on noisy 2D pose input during inference.
    }
    \label{fig:noisy_gen}
    \vspace{-1.2em}
\end{figure}

\section{Conclusion}
\vspace{-0.4em}
In this work, we address the challenging problem of generating fine-grained human action videos that involve significant body deformations and dynamic temporal changes.
To address this issue, 
we propose \textit{FinePhys}, a physics-informed framework that fully explores skeletal data as structural guidance.
The core innovation lies in its comprehensive incorporation of physics through observational, inductive, and learning biases, \textit{i.e.}, by employing in-context learning for dimension lifting, by embedding the Euler-Lagrange equations within neural network modules, and by using appropriate loss functions.
All these ensure the biophysical consistency and motion plausibility of the generated outputs. 
Both quantitative and qualitative results demonstrate FinePhys's superior performance.

\section{Acknowledgments}
This work was founded by the National Natural Science Foundation of China (NSFC) under Grant 62306239, and was also supported by National Key Lab of Unmanned Aerial Vehicle Technology under Grant WR202413.
{
    \small
    \bibliographystyle{ieeenat_fullname}
    \bibliography{main}
}

\clearpage
\maketitlesupplementary

\appendix
\setcounter{tocdepth}{0}
\tableofcontents
\addtocontents{toc}{\setcounter{tocdepth}{2}}

\makeatletter
\renewcommand\subsubsection{\@startsection{subsubsection}{3}{\z@}%
                                     {-3.25ex\@plus -1ex \@minus .2ex}%
                                     {-1em}%
                                     {\normalfont\normalsize\bfseries}}
\makeatother

\section{Training \& Dataset Details}
\subsection{Overview}
We deploy FinePhys using PyTorch, and the training process consists of four steps: \ding{182} Pre-training the skeletal heatmap encoder on the HumanArt~\cite{ju2023humansd} dataset; 
\ding{183} Pre-training the 2D-to-3D module and the PhysNet module on Human3.6M~\cite{ionescu2013human3} and AMASS~\cite{mahmood2019amass} datasets; 
\ding{184} Fine-tune the 2D projection module and PhysNet module using the online detected 2D skeletons detected from FineGym~\cite{shao2020finegym}; 
\ding{185} Jointly fine-tuning the U-Net~\cite{ronneberger2015u}, PhysNet, and 2D projection modules on FineGym. The first three steps of training are conducted on a Linux (Ubuntu) machine with 4 Nvidia 4090 GPUs within 48 hours, while step 4 utilizes two NVIDIA L20 GPUs and completes within 12 hours.

Across all experiments, we apply a linear noise scheduler with 1,000 timesteps, linearly increasing the beta values from 0.00085 to 0.012 to progressively reduce noise during training.
The U-Net backbone incorporates a motion module featuring temporal self-attention layers and positional encoding operating at resolutions \([1, 2, 4, 8]\), enabling multi-scale temporal dynamics capture. 
The motion module is configured with eight attention heads, a single transformer block, and dual temporal self-attention layers to effectively model temporal dependencies. 
To stabilize training, the module parameters are zero-initialized.
We incorporate a Low-Rank Adaptation (LoRA)~\cite{hu2021lora} module with a rank of 64 and a dropout rate of 0.1, facilitating efficient adaptation of the model's spatial and temporal layers while minimizing the number of trainable parameters. Training utilizes the Adam optimizer with an initial learning rate of \(5 \times 10^{-4}\) and a weight decay of \(1 \times 10^{-2}\). Additionally, gradient checkpointing is enabled to optimize GPU memory usage during training.

\subsection{HumanArt Pre-training}
Initially, we train the skeletal heatmap encoder on the HumanArt dataset, a large-scale image collection containing 50K images with accurate pose and text annotations across various scenarios. We leverage the \textit{real-human} subset, comprising 8,750 images with corresponding 2D skeleton annotations. The original COCO-format skeletons are converted to the Human3.6M format, both with 17 keypoints, and subsequently processed into limb heatmaps following the PoseConv3D approach~\cite{duan2022revisiting}. We employ Stable Diffusion v1.5~\cite{rombach2022high} as the spatial generator and keep it frozen during training.

\subsection{Human3.6M and AMASS Pre-training}
To pre-train the 2D-to-3D module and PhysNet, we utilize diverse and realistic 3D human motion data from the Human3.6M and AMASS datasets. Both provide 3D pose annotations essential for skeleton modeling. We use 2D-3D skeleton pairs from Human3.6M as prompt pairs and pre-train both modules for 10 epochs.

\subsection{FineGym Fine-tuning}
\begin{figure*}[t]
    \centering
    \vspace{-1em}
    \includegraphics[width=1.0\linewidth]{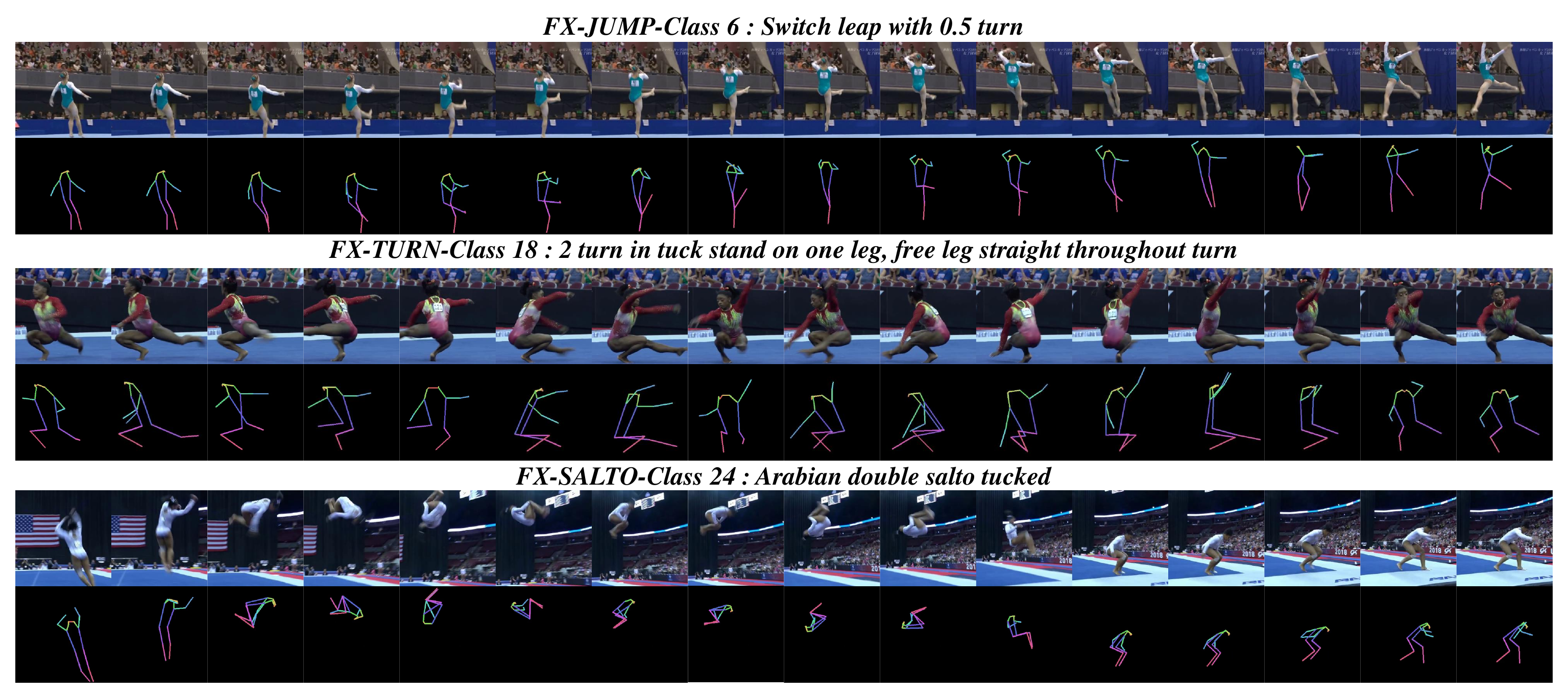}
    \vspace{-1.8em}
    \caption{\textbf{Example videos from FX-JUMP, FX-TURN and FX-SALTO.}
    Each sample video has 16 frames, and the corresponding 2D skeleton sequence is also represented.
    }
    \label{supp_fig:finegym_video}
    \vspace{-1em}
\end{figure*}

For fine-tuning FinePhys, we use the FineGym~\cite{shao2020finegym} dataset, selecting three subsets with distinct motion dynamics: FX-JUMP, FX-TURN, and FX-SALTO. FX-JUMP includes 11 classes (IDs 6--16), FX-TURN comprises 7 classes (IDs 17--23), and FX-SALTO contains 17 classes (IDs 24--40), as detailed in Tab.~\ref{supp_tab:fx-combined}. Example videos and poses are illustrated in Fig.~\ref{supp_fig:finegym_video}.

We generate captions for each video by prompting GPT-4 \cite{achiam2023gpt} to transform existing textual descriptions into standardized prompts.
The instruction provided to GPT-4 was: \textit{``For each gymnastics move described in the labels below, write a detailed description as if explaining to someone who is unfamiliar with gymnastics.''}
 For example, the label \textit{``2 turns on one leg with free leg optional below horizontal''} is converted to \textit{``A person executes two complete turns while balancing on one leg, allowing the lifted leg to remain below hip level or in any chosen position beneath the horizontal line throughout the turning sequence.''}
 This augmentation enhances the model's comprehension of textual prompts, facilitating subsequent video generation tasks.

With the dataset augmented by extended descriptions, we first fine-tune the PhysNet and 2D projection modules for 10,000 training steps using online-detected 2D skeletons from FineGym. Subsequently, we jointly fine-tune the U-Net, PhysNet, and 2D projection modules for an additional 8,000 training steps.

\begin{table}[ht]
\centering
\small
\caption{Categories of FX-JUMP, FX-TURN, and FX-SALTO from Gym99.}
\vspace{-0.8em}
\label{supp_tab:fx-combined}
\renewcommand{\arraystretch}{.9}
\begin{tabularx}{\linewidth}{c|c|X} 
\toprule
\multicolumn{3}{c}{FX-JUMP from Gym99} \\
\midrule
Class & ID & Category \\
\midrule
6 & 0  & Switch leap with 0.5 turn \\
7 & 1  & Switch leap with 1 turn \\
8 & 2  & Split leap with 1 turn \\
 9 & 3  & Split leap with 1.5 turn or more \\
 10 & 4  & Switch leap (leap forward with leg change to cross split) \\
 11 & 5  & Split jump with 1 turn \\
 12 & 6  & Split jump (leg separation 180 degree parallel to the floor) \\
 13 & 7  & Johnson with additional 0.5 turn \\
 14 & 8  & Straddle pike or side split jump with 1 turn \\
 15 & 9  & Switch leap to ring position \\
 16 & 10 & Stag jump \\
\midrule
\multicolumn{3}{c}{FX-TURN from Gym99} \\
\midrule
Class & ID & Category \\
\midrule
 17 & 0  & 2 turn with free leg held upward in 180 split position throughout turn \\
 18 & 1  & 2 turn in tuck stand on one leg, free leg straight throughout turn \\
 19 & 2  & 3 turn on one leg, free leg optional below horizontal \\
 20 & 3  & 2 turn on one leg, free leg optional below horizontal \\
 21 & 4  & 1 turn on one leg, free leg optional below horizontal \\
 22 & 5  & 2 turn or more with heel of free leg forward at horizontal throughout turn \\
 23 & 6  & 1 turn with heel of free leg forward at horizontal throughout turn \\
\midrule
\multicolumn{3}{c}{FX-SALTO from Gym99} \\
\midrule
Class & ID & Category \\
\midrule
 24 & 0  & Arabian double salto tucked \\
 25 & 1  & Salto forward tucked \\
 26 & 2  & Aerial walkover forward \\
 27 & 3  & Salto forward stretched with 2 twist \\
 28 & 4  & Salto forward stretched with 1 twist \\
 29 & 5  & Salto forward stretched with 1.5 twist \\
 30 & 6  & Salto forward stretched, feet land together \\
 31 & 7  & Double salto backward stretched \\
 32 & 8  & Salto backward stretched with 3 twist \\
 33 & 9  & Salto backward stretched with 2 twist \\
 34 & 10 & Salto backward stretched with 2.5 twist \\
 35 & 11 & Salto backward stretched with 1.5 twist \\
 36 & 12 & Double salto backward tucked with 2 twist \\
 37 & 13 & Double salto backward tucked with 1 twist \\
 38 & 14 & Double salto backward tucked \\
 39 & 15 & Double salto backward piked with 1 twist \\
 40 & 16 & Double salto backward piked \\
\bottomrule
\end{tabularx}
\vspace{-1em}
\end{table}


\vspace{-0.5em}
\section{Elaboration on Evaluation Metrics}
In this section, we elaborate on the details of evaluation metrics used in our project.
First, we discuss the limitation of the original CLIP-SIM metric~\cite{radford2021learning} and the corresponding improved CLIP-SIM*.
Then we introduce the details of the user study as well as other metrics.

\vspace{-0.5em}
\subsection{CLIP-SIM Metrics and Limitations}
We analyze the CLIP-SIM metric based on three aspects: \textit{semantic consistency}, \textit{domain consistency}, and \textit{temporal consistency}~\cite{guo2023animatediff}. 
Below, we detail each aspect and discuss their limitations.

\vspace{-1.5em}
\subsubsection*{\ding{182} \textit{Semantic Consistency}}
measures the alignment between textual prompts and the generated video frames. Specifically, for a given text prompt \( P \) and a generated video \( \Tilde{V} \) with \( T \) frames, the semantic consistency score is computed as the average CLIP similarity between \( P \) and each frame of \( \Tilde{V} \):
\begin{align}
    \text{CLIP}_{\text{text}} (P, \Tilde{V}) = \frac{1}{T}\sum_{t=1}^T \text{CLIP} (P, \Tilde{V}(t)).
\end{align}

\noindent \textbf{Limitations of $\text{CLIP}_{\text{text}}$:} The original semantic consistency metric struggles with fine-grained action labels due to semantic ambiguity and entanglement in the CLIP embedding space. As illustrated in Fig.~\ref{supp_fig:metric_text}, while the metric performs adequately for coarse-grained action categories (\textit{e.g.}, those from UCF101~\cite{soomro2012ucf101}), it fails with FineGym labels where the embedded vectors of specific categories overlap significantly, rendering the metric ineffective for distinguishing between similar fine-grained actions.

\vspace{-1em}
\subsubsection*{\ding{183} \textit{Domain Consistency}}

assesses the similarity between generated video frames and reference images generated by an open-sourced image generation model, such as Stable Diffusion~\cite{rombach2022high}. For a reference image \( I \) and a generated video \( \Tilde{V} \) with \( T \) frames, the domain consistency score is calculated as:
\begin{align}
    \text{CLIP}_{\text{domain}} (I, \Tilde{V}) = \frac{1}{T}\sum_{t=1}^T \text{CLIP} (I, \Tilde{V}(t)).
\end{align}

\begin{figure}[t]
    \centering
    \includegraphics[width=1.0\linewidth]{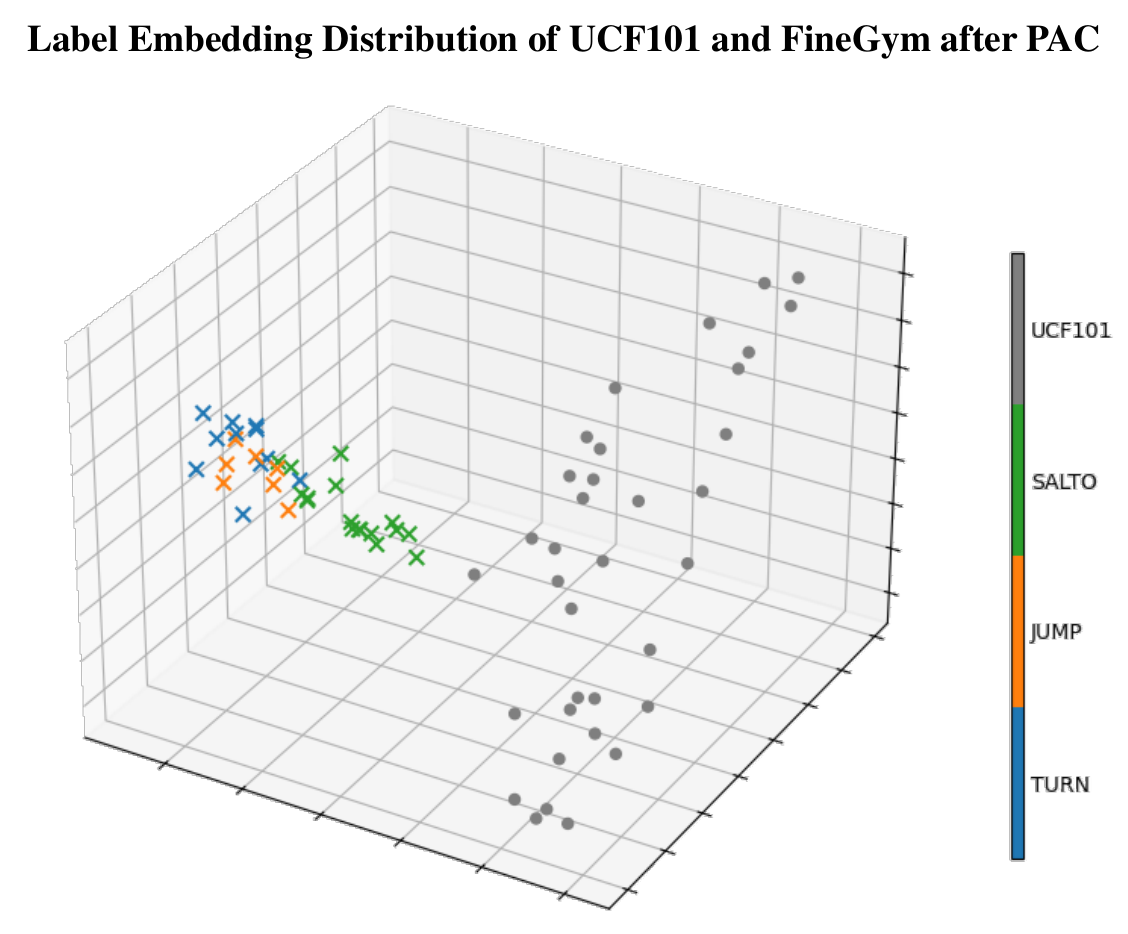}
    \vspace{-1.8em}
    \caption{\textbf{Limitations of semantic consistency in original CLIP-SIM.} 
    We utilize CLIP models to obtain the embedded textual features and Probably Approximately Correct (PAC) for dimensionality reduction.
    The distribution of embedded category labels from FX-JUMP, FX-TURN and FX-SALTO as well as UCF101 is shown. Label features from FineGym are entangled, while those from UCF101 are clearly seperated.
    }
    \label{supp_fig:metric_text}
    \vspace{-1em}
\end{figure}

\noindent \textbf{Limitations of $\text{CLIP}_{\text{domain}}$:} The domain consistency metric is unreliable for fine-grained actions because reference images generated by Stable Diffusion may not accurately reflect the nuances of specific actions or their dynamics, as shown in Fig.~\ref{supp_fig:metric_domain}.
Additionally, comparing the generated results in Fig.\ref{supp_fig:fig_C_3_turn}, higher domain scores do not necessarily correspond to better representations of fine-grained videos. 
For instance, T2V-Zero generates nonsensical content that still achieves a higher domain score than AnimateDiff, and VideoCrafter's highest-scoring results often contain visible artifacts and limb inaccuracies.

\vspace{-1em}
\subsubsection*{\ding{184} \textit{Temporal Consistency}}
evaluates the smoothness of transitions between frames in a generated video by computing the average CLIP similarity between randomly selected pairs of frames. Given a generated video \( \Tilde{V} \) and a set of \( N \) frame pairs \( \mathbb{P} \), the temporal consistency score is:
\begin{align}
    \text{CLIP}_{\text{smooth}} (\Tilde{V}) = \frac{1}{N}\sum_{(i, j)\in\mathbb{P}} \text{CLIP} (\Tilde{V}(i), \Tilde{V}(j)).
\end{align}

\noindent \textbf{Limitations of $\text{CLIP}_{\text{smooth}}$:} The original temporal consistency metric is unsuitable for fine-grained human actions, which inherently involve rapid and significant temporal changes. As demonstrated in Fig.\ref{supp_fig:fig_C_3_jump}, models like T2I-Zero that generate predominantly static scenes paradoxically achieve the highest temporal consistency scores. This indicates that the metric fails to capture the dynamic nature of fine-grained actions, instead rewarding unnaturally smooth or static video sequences.

\subsection{The Improved CLIP-SIM* Metrics}
To overcome the aforementioned limitations, we propose an enhanced version of CLIP-SIM, termed CLIP-SIM*, specifically designed for evaluating fine-grained human action videos. CLIP-SIM* refines the calculations of domain consistency and temporal consistency by adopting a data-driven approach, while leaving the original semantic consistency as a minor metric.

\begin{figure}[t]
    \centering
    \vspace{-1em}
    \includegraphics[width=1.0\linewidth]{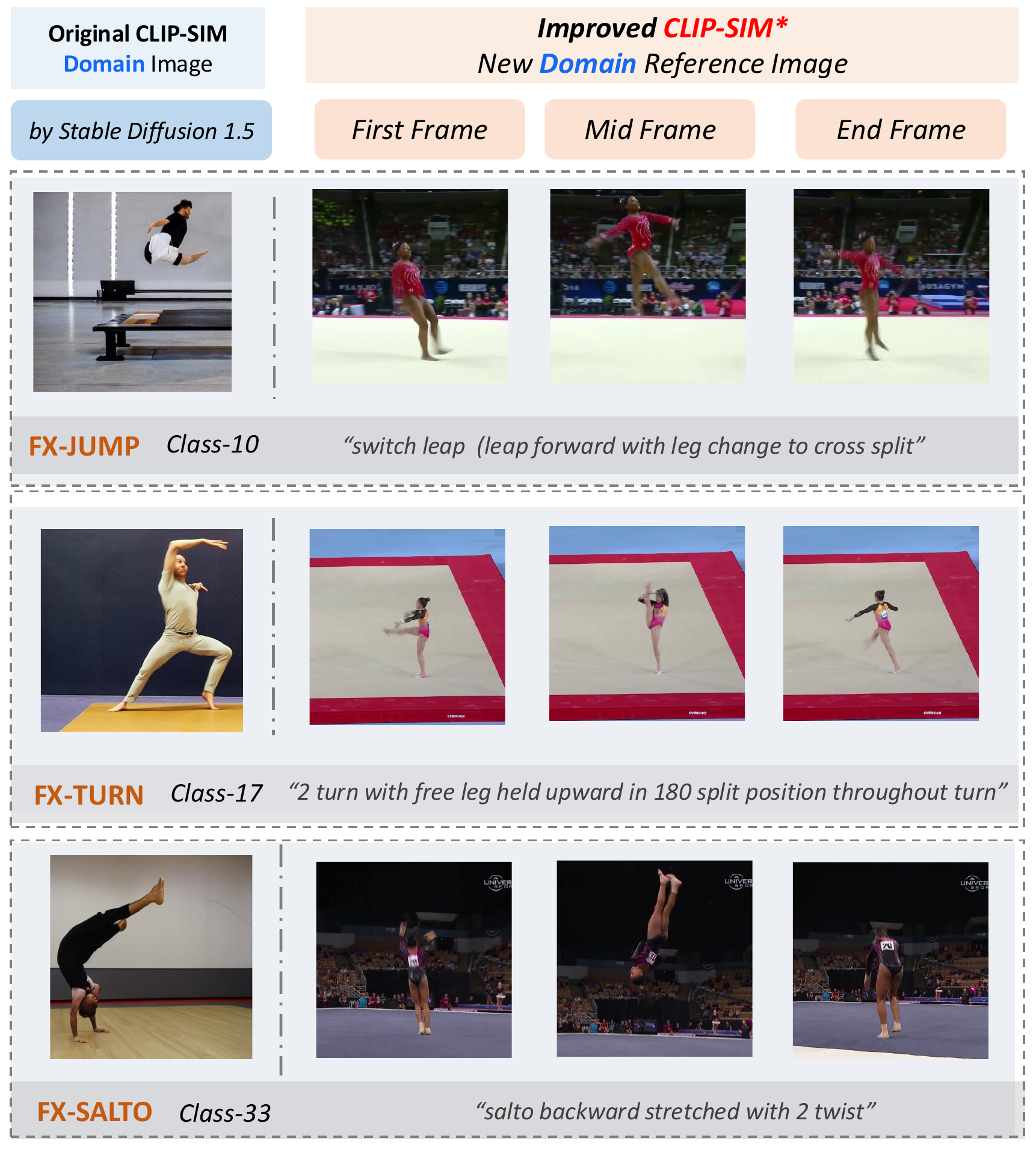}
    \vspace{-1.8em}
    \caption{\textbf{Domain image of original CLIP-SIM and the improved CLIP-SIM* from FX-JUMP, FX-TURN and FX-SALTO.} Reference images generated by Stable Diffusion may not accurately reflect the nuances of specific actions or their dynamics (Original CLIP-SIM), while CLIP-SIM* randomly selects one video from the given class and extracts three representative frames (start, middle, end) to form a more reasonable reference set.
    }
    \label{supp_fig:metric_domain}
    \vspace{-1em}
\end{figure}

\vspace{-1em}
\subsubsection*{\ding{182} \textit{Improved Domain Consistency}.}
 Instead of relying on reference images generated by Stable Diffusion, CLIP-SIM* leverages ground-truth videos to select more relevant reference images. 
 Specifically, we randomly choose ground-truth videos and extract three representative frames (\textit{start, middle, end}) from each to form the reference set \( \{I_j\}_{j=1}^{N} \), as shown in the right part of Fig.~\ref{supp_fig:metric_domain}. 
 
 The domain consistency score is then computed as the average CLIP similarity between each generated frame and all reference images:
\begin{align}
    \text{CLIP}_{\text{text}}^* (\Tilde{V}, \{I_j\}) = \frac{1}{N} \cdot \frac{1}{T} \sum_{t=1}^T \sum_{j=1}^N \text{CLIP} (\Tilde{V}(t), I_j).
\end{align}
This approach ensures that the reference images are contextually and semantically aligned with the fine-grained actions being evaluated, thereby providing a more accurate measure of domain consistency.

\vspace{-1em}
\subsubsection*{\ding{183} \textit{Improved Temporal Consistency}.}
To better assess the temporal dynamics of fine-grained actions, we propose an improved temporal consistency metric within CLIP-SIM*, which preserves the temporal changing patterns inherent to specific action classes. Instead of enforcing smoothness across all frames, CLIP-SIM* compares the generated video with multiple reference videos from the same action category. For each action label, we select \( M \) reference videos \( V^{\text{Ref}} \) and uniformly sample \( K_i \) frames from each reference video, where \( K_i \in \{1, 2, 4, 8, 16\} \). The temporal consistency score is then calculated as:
\begin{align}
    \text{CLIP}_{\text{smooth}}^*(\Tilde{V}, V^{\text{Ref}}) = \sum_{l=1}^M \sum_{k=1}^{K_i} \text{CLIP} (\Tilde{V}(k), V^{\text{Ref}}_l(k)).
\end{align}
This modification allows $\text{CLIP}_{\text{smooth}}^*$ to effectively measure whether the generated video replicates the temporal dynamics of specific fine-grained actions, addressing the shortcomings of the original temporal consistency metric, as shown in Fig.\ref{supp_fig:fig_C_3_salto}.

\subsection{Details of User Study}

As discussed in the main paper, we evaluate the generation results through a user study, which provides a more reliable assessment. In practice, each participant is presented with a series of text-video, image-video, and video-video pairs and asked to rate \textit{semantic consistency}, \textit{temporal consistency}, and \textit{domain consistency} on a scale \textbf{from 1 to 5}. The layout of the user study interface is illustrated in Fig.~\ref{supp_fig:user_study}.

Specifically, we developed a questionnaire that tested all baseline models alongside our results. Each video result was accompanied by the same textual descriptions, reference images, and reference videos. Participants were instructed to objectively evaluate the similarity of the video results to this reference information. To ensure impartiality, we omitted any details about the models used and distributed the questionnaire to 20 professionals unfamiliar with our work, thereby obtaining objective data.

\begin{figure*}[t]
    \centering
    \vspace{-1em}
    \includegraphics[width=1.0\linewidth]{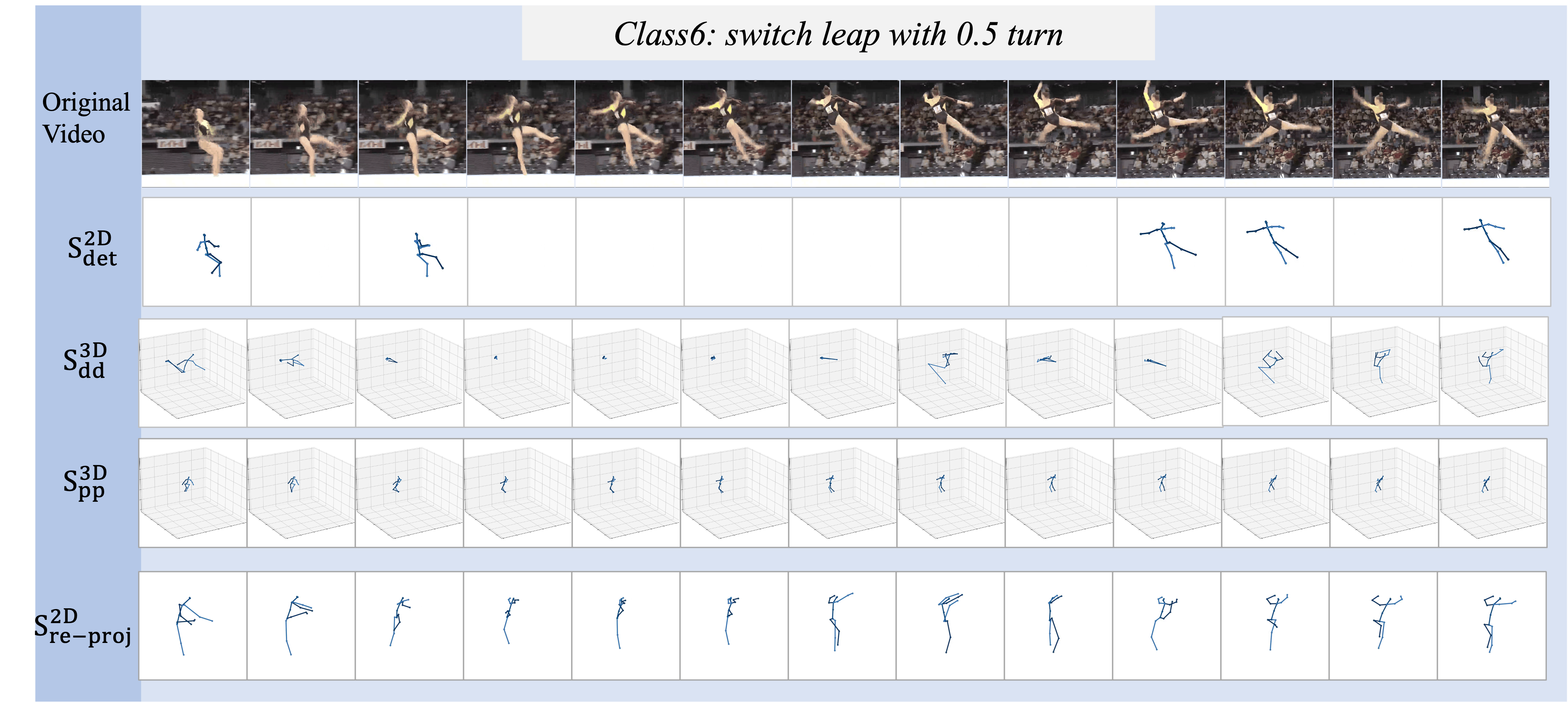}
    \vspace{-1.8em}
    \caption{\textbf{Visualization of different pose sequences} on the class \textit{``switch leap with 0.5 turn"} from the \textbf{FX-Jump} subset, demonstrating the complete transformation process within our framework.
    }
    \label{supp_fig:fig_c_2_jump}
\end{figure*}

\subsection{Other Metrics}
\subsubsection*{PickScore.}
PickScore~\cite{kirstain2023pick} trains a scoring function \( s(\cdot) \) based on the CLIP framework using the large-scale user preferences dataset Pick-a-Pic to score the quality of generated images. Its performance in assessing generated images surpasses that of other evaluation metrics, even outperforming expert human annotators.

Given a text prompt \( P \) and an image \( I \) as input, PickScore calculates the score of the generated image as follows:
\begin{align}
    s(P, I) = E_{\text{txt}}(P) \cdot E_{\text{img}}(I) \cdot \tau
\end{align}
where \( E_{\text{txt}} \) and \( E_{\text{img}} \) represent the text encoder and image encoder, respectively, and \( \tau \) denotes the learned scalar temperature parameter of CLIP.

While PickScore was originally developed for image evaluation, we have extended it to the domain of video evaluation. Specifically, given a text prompt \( P \) and a generated video \( \Tilde{V} \), we compute the average PickScore across all frames of the video:
\begin{align}
    \text{PickScore}(P, \Tilde{V}) = \frac{1}{T} \sum_{t=1}^T s(P, \Tilde{V}(t))
\end{align}
where \( \Tilde{V}(t) \) denotes the \( t \)-th frame of the generated video, and \( T \) is the total number of frames.

\vspace{-1em}
\subsubsection*{Fr\'{e}chet Video Distance (FVD).}
FVD~\cite{unterthiner2018towards} is a widely used metric for evaluating video generation models. In the domain of temporal analysis \cite{deng2024disentangling, deng2025parsimony}, it is highly correlated with the visual quality of generated samples and assesses temporal consistency. FVD utilizes a pre-trained video recognition model to extract features from both real and generated videos, forming two sets of features, and then computes the mean and covariance matrices of these two sets. The FVD is represented as the Fr\'{e}chet distance between these two distributions:
\begin{align}
    \text{FVD} = \lVert \mu - \tilde{\mu} \rVert^2 + \text{Tr}(\Sigma + \tilde{\Sigma} - 2(\Sigma \tilde{\Sigma})^{\frac{1}{2}})
\end{align}
where $\mu$ and $\Sigma$ are the mean and covariance matrix of the real video feature set, while $\tilde{\mu}$ and $\tilde{\Sigma}$ are the mean and covariance matrix of the generated video feature set.
However, as observed in~\cite{kwon2024harivo}, unsatisfactory video generation results could achieve a higher FVD score, challenging its reliability.


\section{Additional Illustration \& Analysis}
\subsection{Elaboration on Euler-Lagrange Equations}
In the main paper, we use the following equation to represent the process in Lagrangian Mechanics:
\begin{align}
    M(q) \ddot{q} = J(q, \dot{q}) - C (q, \dot{q}),
\end{align}
which is a common form used in robotics and dynamics, known as the equation of motion in terms of mass matrix $ M(q)$, generalized forces $J(q, \dot{q})$, and Coriolis and centrifugal forces $C (q, \dot{q})$.
Here we elaborate on its relation with the original Euler-Lagrange Equations, \textit{i.e.}:
\begin{align}
    \frac{\partial L}{\partial q^i}(t, q(t), \dot{q}(t)) - \frac{d}{dt}\frac{\partial L}{\partial \dot{q}^i} (t, q(t), \dot{q}(t)) = 0.
\end{align}
Assume the kinetic energy of the system is given by $T=\frac{1}{2} \dot{q}^T M(q) \dot{q}$, 
and the potential energy is typically a function of the generalized coordinates $q$ denoted by $V = V(q)$,
then the Lagrangian is defined as:
\begin{align}
L = T - V = \frac{1}{2} \dot{q}^T M(q) \dot{q} - V(q).
\end{align}
Then we calculate $\frac{\partial L}{\partial q^i}$ and $\frac{\partial L}{\partial \dot{q}^i}$:
\begin{align}
    \frac{\partial L}{\partial q^i} &= -\frac{\partial V}{\partial q^i} + \frac{1}{2} \dot{q}^T \frac{\partial M(q)}{\partial q^i} \dot{q} \\
    \frac{d}{dt} \left( M_{ij}(q) \dot{q}^j \right) &= \dot{q}^j \frac{\partial M_{ij}}{\partial q^k} \dot{q}^k + M_{ij}(q) \ddot{q}^j
\end{align}
and substitute these results into the Euler-Lagrange equation:
\begin{align}
    -\frac{\partial V}{\partial q^i} + \frac{1}{2} \dot{q}^T \frac{\partial M(q)}{\partial q^i} \dot{q} - \left( \dot{q}^j \frac{\partial M_{ij}}{\partial q^k} \dot{q}^k + M_{ij}(q) \ddot{q}^j \right) = 0,
\end{align}
where $-\frac{\partial V}{\partial q^i}$ represents the partial derivative of the potential energy with respect to the coordinates, i.e., the generalized force, i.e., $J(q, \dot{q})$.
Thus we could obtain the following formulation:
\begin{align}
    M(q)\ddot{q} = J(q, \dot{q}) - C(q, \dot{q}).
\end{align}

\subsection{Visualization of the Pose Modality}

\begin{figure*}[t]
    \centering
    \includegraphics[width=1.0\linewidth]{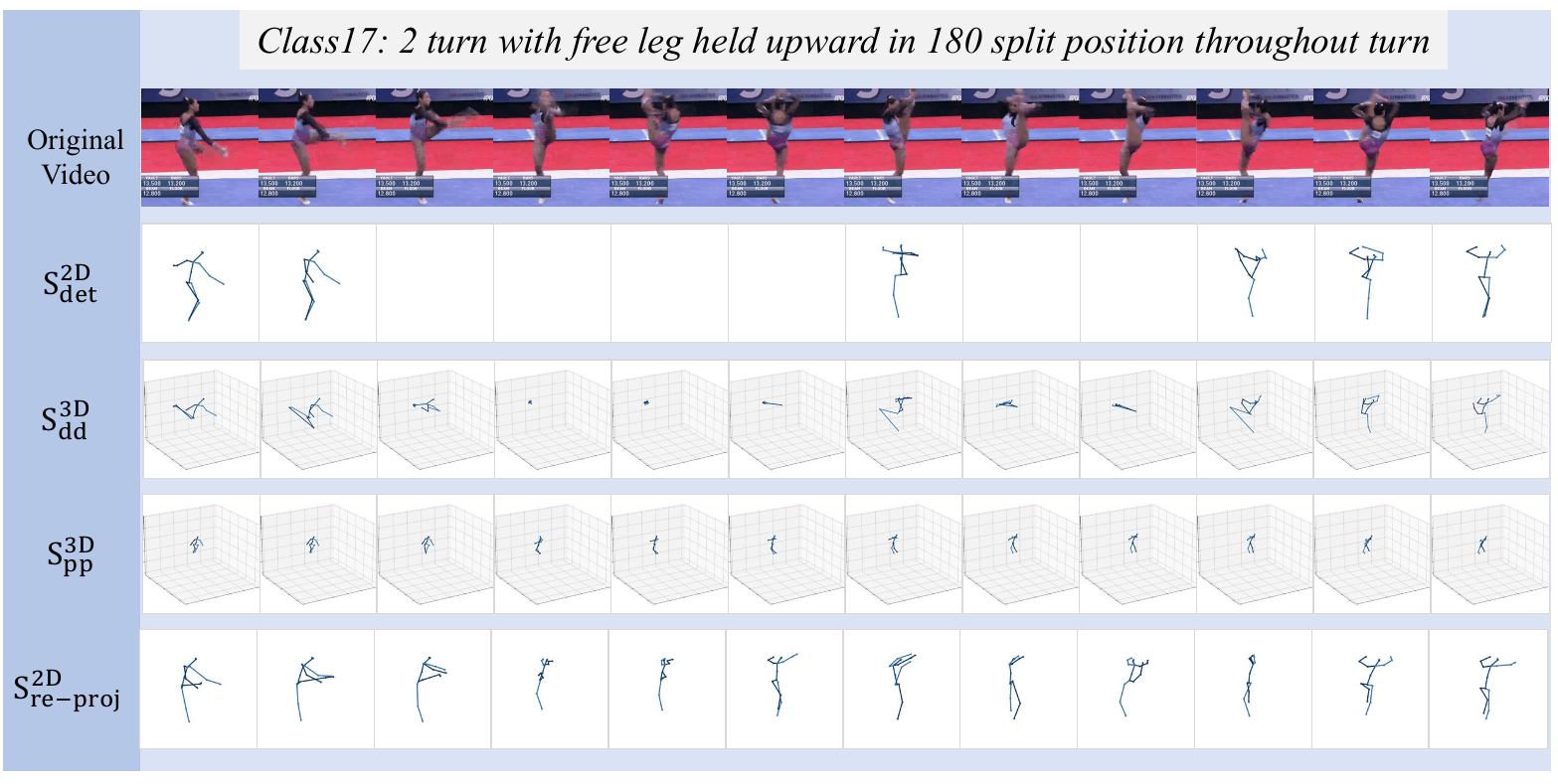}
    \vspace{-1.8em}
    \caption{\textbf{Visualization of different pose sequences} on the class \textit{``2 turn with free leg held upward in 180 split position throughout turn"} from the \textbf{FX-Turn} subset, demonstrating the complete transformation process within our framework.
    }
    \label{supp_fig:fig_c_2_turn}
    \vspace{-0.8em}
\end{figure*}

Recall that our FinePhys framework fully leverages skeletal data through a sequence of specialized modules:
(1) The online pose estimator generates detected 2D poses, denoted as \( S_{\text{detect}}^{2D} \);
(2) Then the in-context-learning module processes and transforms them into $S_{dd}^{3D}$;
(3) After the PhysNet module we obtain $S_{pp}^{3D}$,
(4) and finally we re-projected the average of $S_{dd}^{3D}$ and $S_{pp}^{3D}$ into 2D space to obtain  $S_{\text{re-proj}}^{2D}$.

Fig.\ref{supp_fig:fig_c_2_jump}, Fig.~\ref{supp_fig:fig_c_2_turn}, Fig.~\ref{supp_fig:fig_c_2_salto} present additional visualizations of these pose sequences, illustrating the entire transformation process within our framework. 
Due to the large variation and high complexity of fine-grained actions, the detected 2D poses (\( S_{\text{detect}}^{2D} \)) exhibit significant misidentifications across joints throughout the video. The in-context learning module improves these poses, enabling \( S_{dd}^{3D} \) to partially reconstruct missing or distorted skeletons in each frame. 
However, in cases of severe distortion, the data-driven approach becomes unstable, resulting in \( S_{dd}^{3D} \) being noisy and physically implausible. 
The PhysNet module mitigates this issue by producing \( S_{pp}^{3D} \), which is more stable and constrained, effectively correcting deviations in \( S_{dd}^{3D} \). 
Consequently, the averaged and re-projected 2D poses (\( S_{\text{re-proj}}^{2D} \)) show substantial improvements compared to the original detections, validating the efficacy of our approach.

\begin{figure*}[t]
    \centering
    \includegraphics[width=1.0\linewidth]{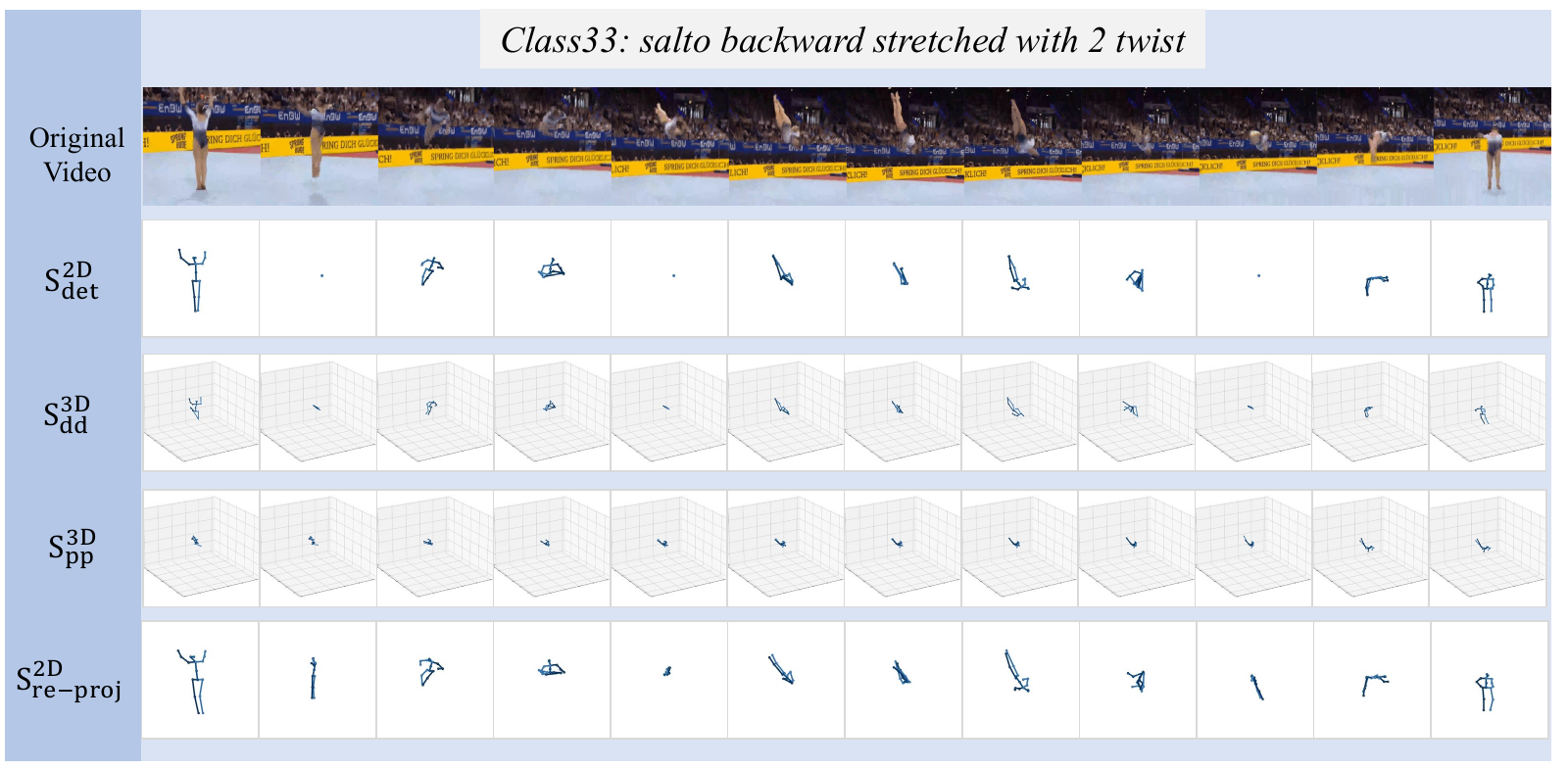}
    \vspace{-1.8em}
    \caption{\textbf{Visualization of different pose sequences} on the class \textit{``salto backward stretched with 2 twist"} from the \textbf{FX-Salto} subset, demonstrating the complete transformation process within our framework.
    }
    \label{supp_fig:fig_c_2_salto}
\end{figure*}

\subsection{More Generated Results and Comparison}

In this section, we present additional qualitative results to demonstrate the effectiveness of our proposed FinePhys framework in generating fine-grained human action videos.

We compare the generated results of FinePhys with those of baseline methods across three action subsets: FX-JUMP, FX-TURN, and FX-SALTO, as illustrated in Fig.~\ref{supp_fig:fig_C_3_jump}, Fig.~\ref{supp_fig:fig_C_3_turn}, and Fig.~\ref{supp_fig:fig_C_3_salto}, respectively. The key observations are as follows:
\ding{182}
Our CLIP-SIM* metric more accurately reflects the quality of video generation compared to the original CLIP-SIM metric. For example, methods such as Follow-Your-Pose and Latte achieve high scores on the original Domain Score, yet the generated actions exhibit significant inconsistencies with physical laws. Similarly, T2V-zero attains the highest score on the Smooth Score by generating continuous identical frames, which lack realistic motion dynamics. In contrast, CLIP-SIM* scores align more closely with human intuition, providing a more reliable assessment of video quality.

\ding{183}
FinePhys consistently outperforms other baseline methods across different action categories. Baseline methods that lack guidance from physical information often produce unrealistic limb movements. For instance, Latte displays multiple limb artifacts in Class 14, and VideoCraft shows unrealistic levitation in Class 20
. In contrast, FinePhys incorporates physics modeling through the PhysNet module, resulting in more natural and coherent actions that adhere to real-world physical constraints.



\subsection{Limitation and Future Work.}

\subsubsection*{Intractable Cases.}

Although FinePhys outperforms its competitors in generating results, significant challenges remain unresolved. High-speed motions and substantial body deformations pose considerable difficulties, particularly when they are intertwined, as seen in \textit{salto} routines. Generating fine-grained actions such as \textit{double salto backward stretched} is currently intractable, as shown in Fig.~\ref{supp_fig:fig_C_4_limitation_1}, let alone accurately distinguishing between actions like \textit{"salto backward stretched with 2.5 twist"}, \textit{"salto backward tucked with 1 twist"}, and \textit{"double salto backward tucked with 1 twist"}. We encourage future research efforts to address these complex scenarios.

\vspace{-1em}
\subsubsection*{Reliance on Initial Pose Detection.}

FinePhys fully utilizes the pose modality; however, the initial step of the pipeline involves online 2D pose estimation. Due to the complexity of fine-grained human actions, we observed that the online pose estimator can occasionally fail completely, resulting in no detected 2D poses, as shown in Fig.~\ref{supp_fig:fig_C_4_limitation_2}. In such cases, the initial poses rely entirely on the pose prior used in the in-context learning module. Even if we can restore the human structure spatially, no motion is present. 
In future work, we will consider selecting appropriate scenarios to evaluate our current FinePhys implementations and explore additional modalities (e.g., optical flow) to address this issue.

\begin{figure}[t]
    \centering
    \includegraphics[width=1.0\linewidth]{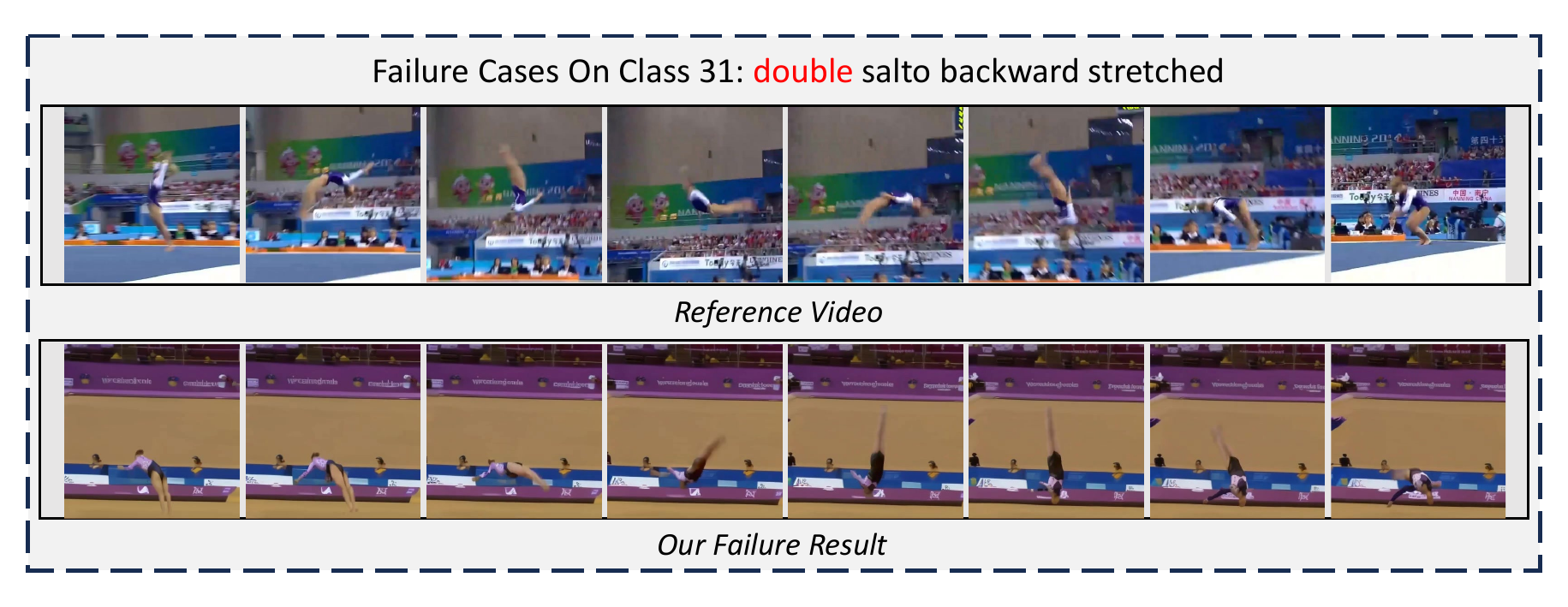}
    \vspace{-1.8em}
    \caption{\textbf{Limitations in intractable cases.} For class 31: \textit{double salto backward stretched}, FinePhys fails to generate a double salto, resulting in only a single flip being observed.
    }
    \label{supp_fig:fig_C_4_limitation_1}
\end{figure}

\begin{figure}[t]
    \centering
    \includegraphics[width=1.0\linewidth]{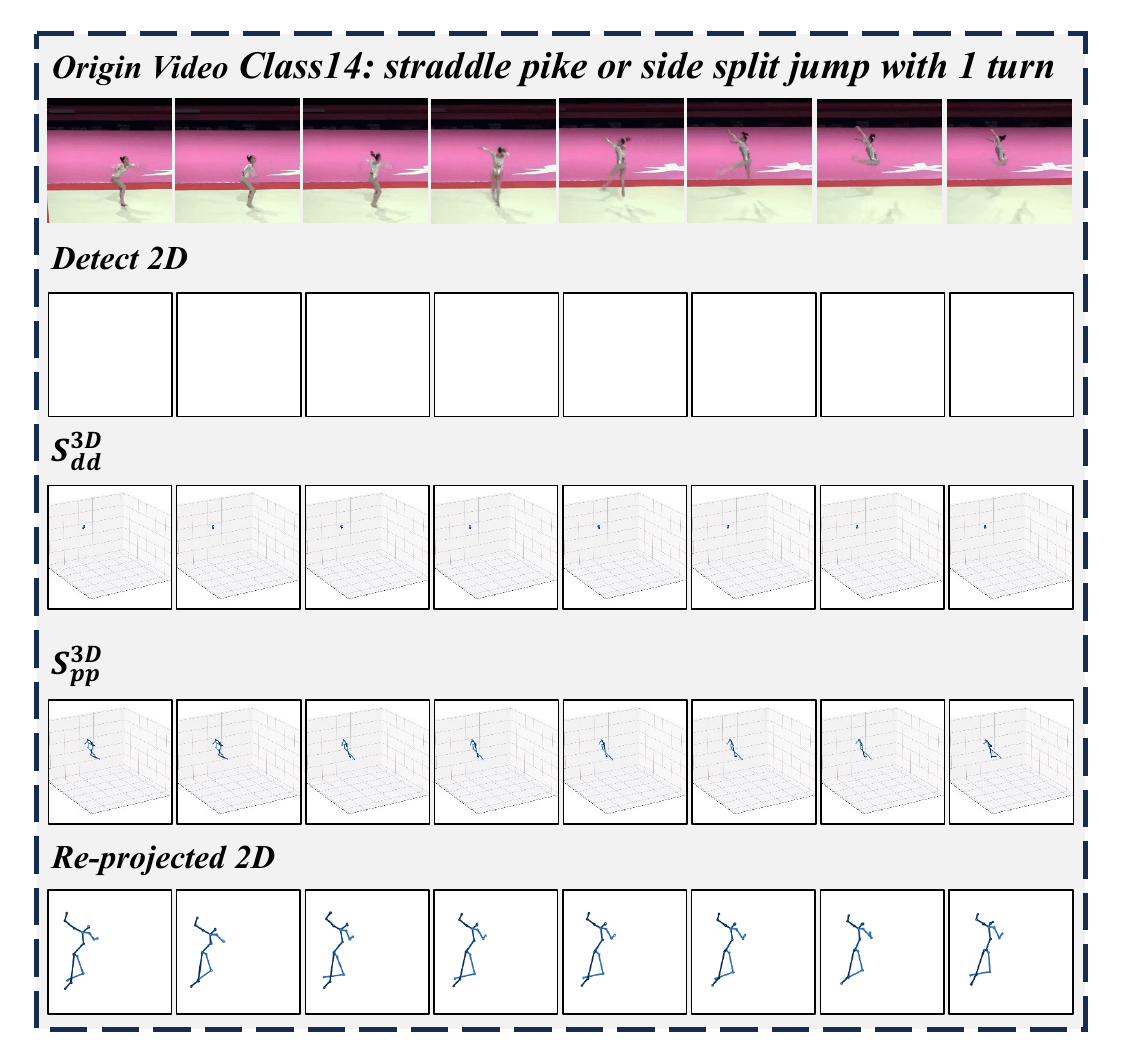}
    \vspace{-1.8em}
    \caption{\textbf{Negative Impact of Initial Pose Detection.}.  Current online pose estimators may fail completely due to the complexity of fine-grained human actions, which affects subsequent processing stages in the FinePhys framework. Even when the physical structure of the human body is spatially restored, the intricate motion dynamics cannot be accurately reconstructed, resulting in unrealistic or static video outputs. 
    }
    \label{supp_fig:fig_C_4_limitation_2}
\end{figure}

\vspace{-1em}
\subsubsection*{Focus on Fine-grained Human Actions.}
Although video generation techniques have been extensively explored and improved, applying these methods to the specific and challenging domain of fine-grained human actions can reveal the limitations of current approaches and inspire future advancements \cite{shao2018find, huang2025vistadpo, shao2020intra, chen2024finecliper}. In this work, we select three fine-grained human action subsets, each encoding distinct motion dynamics: 
 \ding{182} \textit{Turning} Focuses on precise rotational movements; 
\ding{183}\textit{Jumping} emphasizes rapid vertical motion combined with moderate rotations;
\ding{184} While \textit{Salto} involves complex aerial maneuvers with multiple twists and flips, and is the most challenging.
By conducting comprehensive quantitative comparisons alongside qualitative analyses, we aim to draw greater attention to the challenges inherent in generating fine-grained human actions. This focused evaluation not only highlights the strengths and weaknesses of existing methods but also provides valuable insights for future research and development in this domain.

\vspace{-1em}
\subsubsection*{Further Exploration on Physics.} 
In future work, we aim to enhance the integration of physics modeling in video generation from diverse perspectives, such as collision dynamics, fluid interactions, etc. Currently, generating fine-grained human actions restricts the model's ability to focus solely on motion dynamics, as it must also account for the spatial structure of the human body \cite{chen2024gaussianvton, xiu2023econ, xiu2022icon, tang2023predicting}. To address this complexity, we plan to simplify scenes by utilizing basic geometric shapes for environmental interactions, thereby reducing model complexity while maintaining a robust incorporation of physical principles. Additionally, we will investigate the incorporation of physical laws into video generation, which may involve developing new algorithms or refining existing techniques to more accurately simulate real-world physical behaviors.

\begin{figure*}[t]
    \centering
    \vspace{-1em}
    \includegraphics[width=1.0\linewidth]{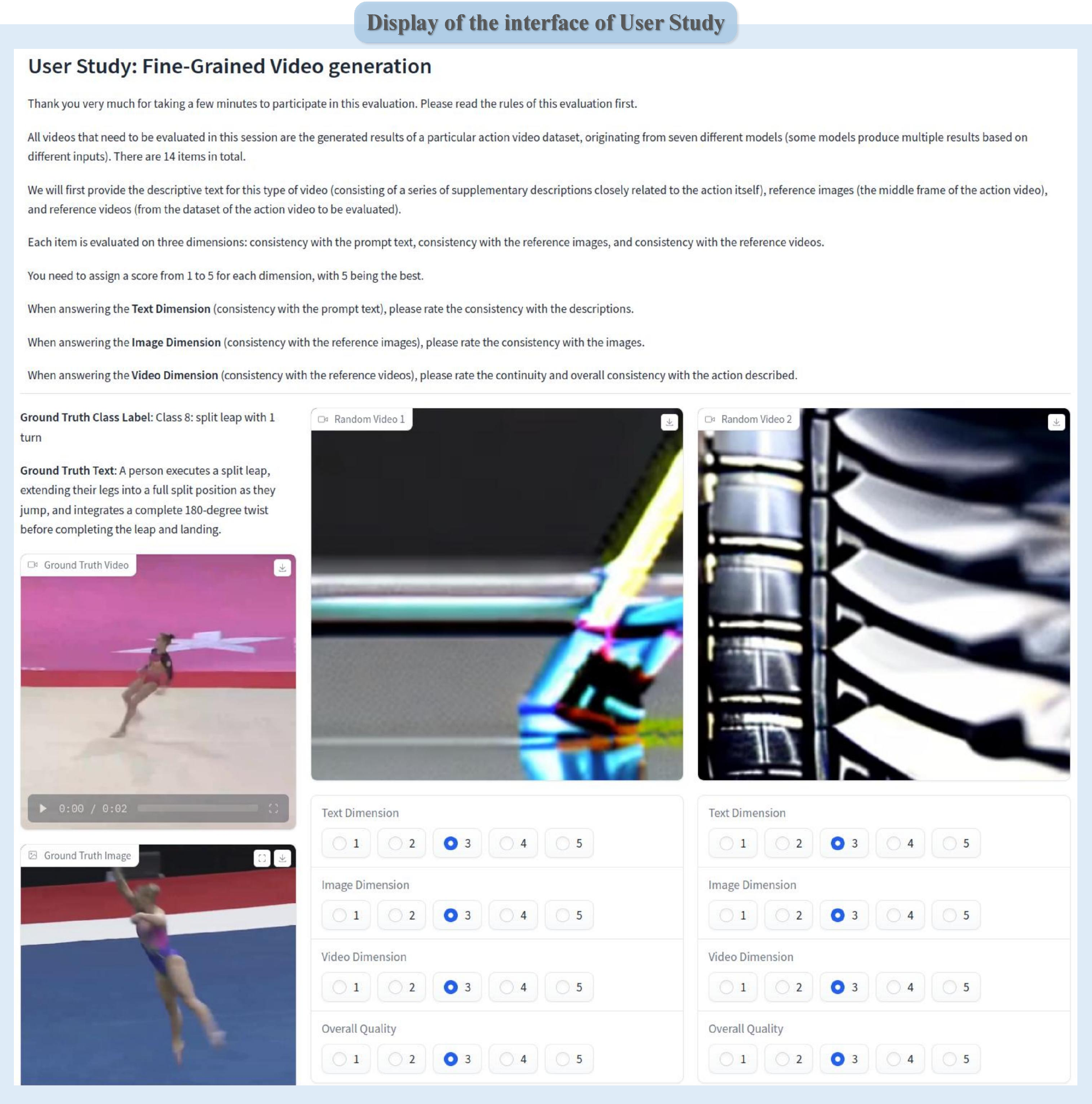}
    \vspace{-1.8em}
    \caption{\textbf{Display of the interface of User Study.}
    }
    \label{supp_fig:user_study}
    \vspace{-1em}
\end{figure*}

\begin{figure*}[t]
    \centering
    \includegraphics[width=1.0\linewidth]{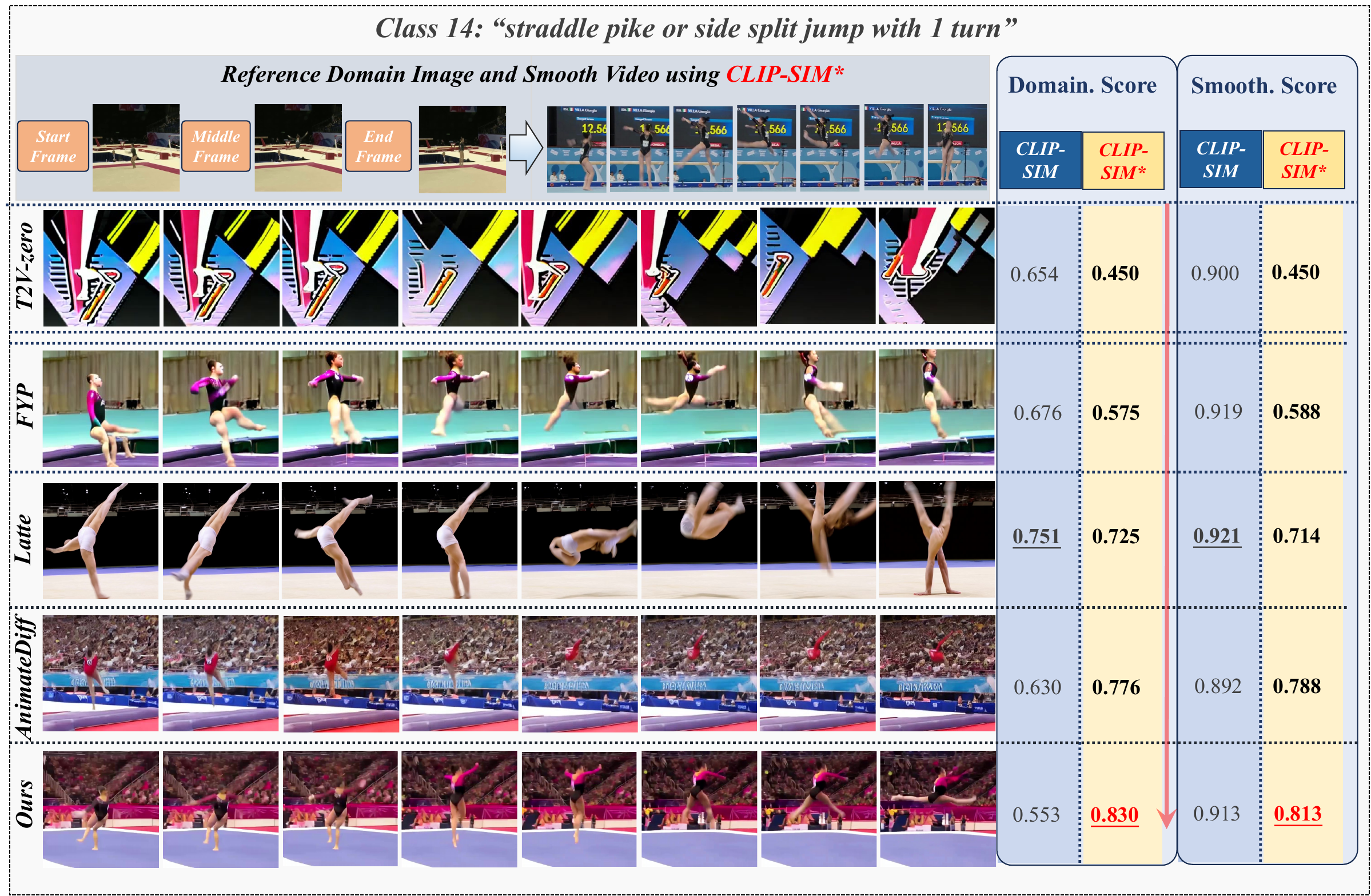}
    \vspace{-1.8em}
    \caption{\textbf{Qualitative Results on FX-JUMP.} 
    FX-JUMP focuses on the motion continuity of the gymnastics' body.
    Compared with other
    baselines, our method demonstrates superior performance in
    understanding physical consistency.
    }
    \label{supp_fig:fig_C_3_jump}
\end{figure*}

\begin{figure*}[t]
    \centering
    \includegraphics[width=1.0\linewidth]{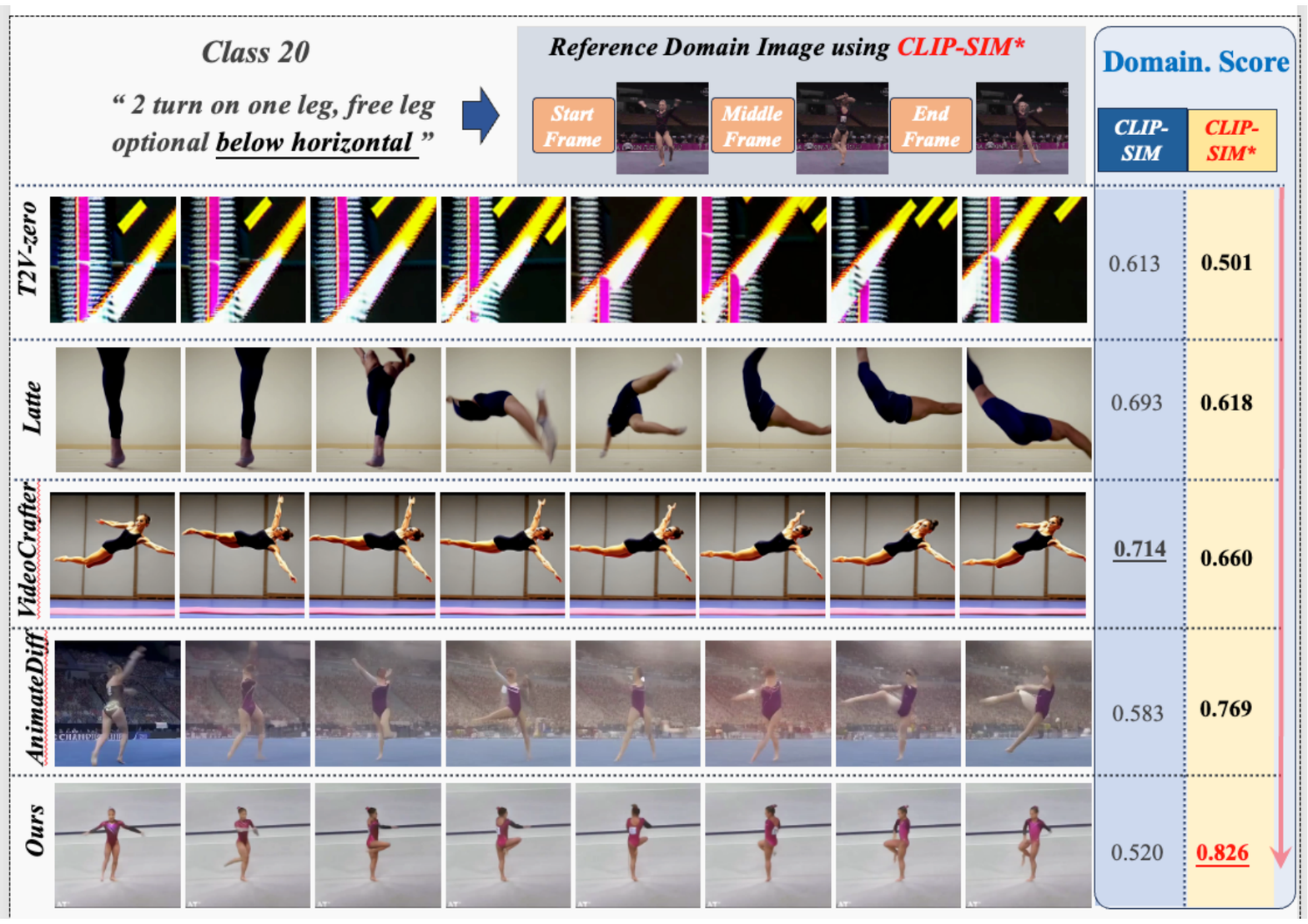}
    \vspace{-1.8em}
    \caption{\textbf{Qualitative Results on FX-TURN.} FX-TURN focuses on the minor difference of the gymnastics' body.
    Compared with other
    baselines, our method demonstrates superior performance in
    understanding complex and fine-grained semantics, keeping
    the consistency of bio-physical characteristics, and adhering
    to the physical principles.
    }
    \label{supp_fig:fig_C_3_turn}
\end{figure*}

\begin{figure*}[t]
    \centering
    \includegraphics[width=1.0\linewidth]{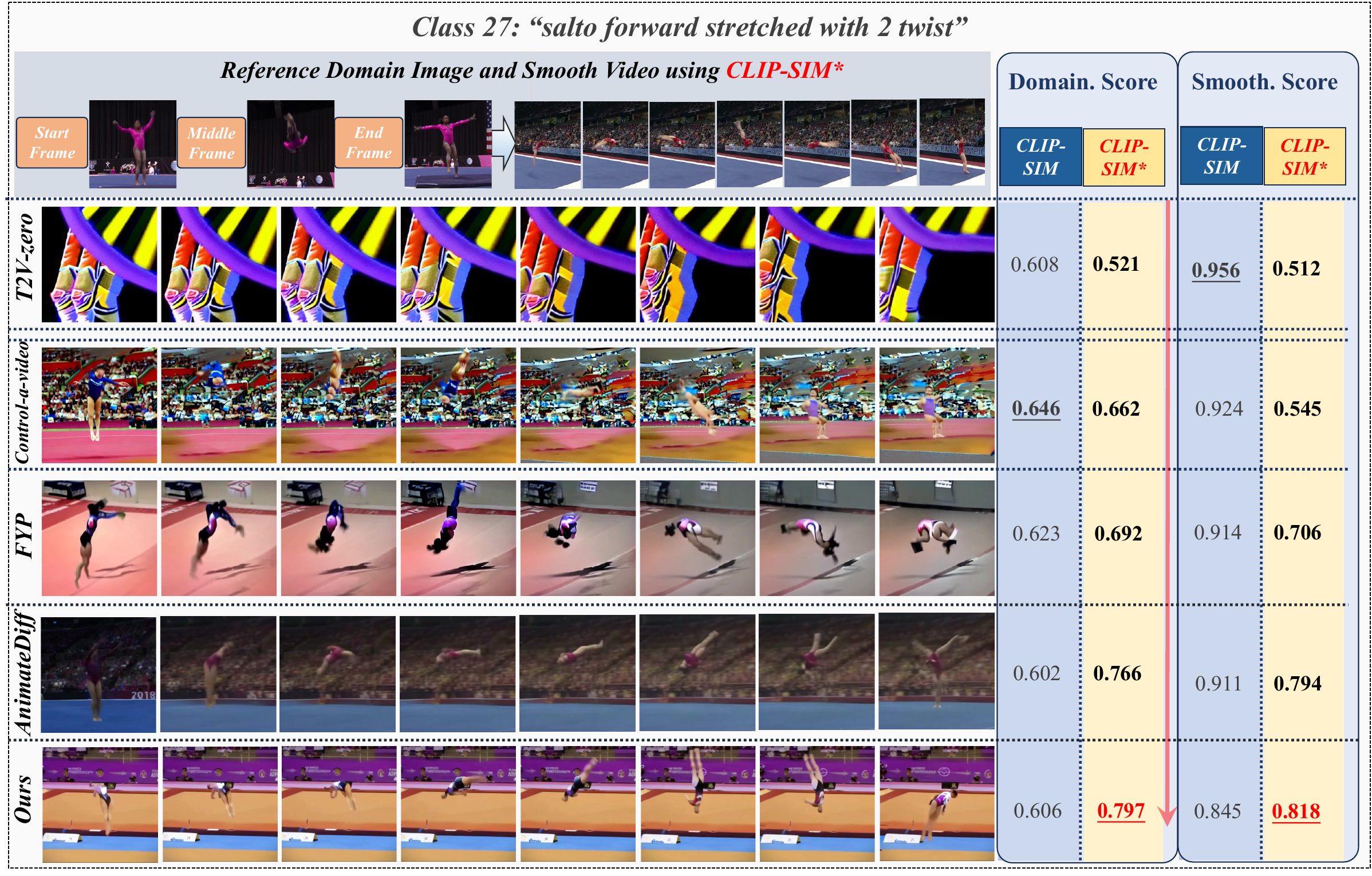}
    \vspace{-1.8em}
    \caption{\textbf{Qualitative Results on FX-SALTO.}  FX-SALTO demands gymnastics’s body rotates 360° around a horizontal axis
    with the feet passing over the head, which is the most difficult in all of three sub-datasets in FineGym. Compared with other
    baselines, results in our methods maintain better temporal consistency, more adhering to the bio-physical rules.
    }
    \label{supp_fig:fig_C_3_salto}
\end{figure*}

\end{document}